\documentclass[a4paper,fleqn]{cas-dc}

\usepackage[numbers]{natbib}

\usepackage{acronym} 
\usepackage{amsmath}
\usepackage{amssymb}
\usepackage{amsthm}
\usepackage{color} 
\usepackage{float}
\usepackage{graphicx} 
\usepackage{multirow} 
\usepackage{stackengine} 
\usepackage{xargs}
\usepackage{tabularx}  
\usepackage{siunitx}
\usepackage{graphicx}    
\usepackage{caption}     
\usepackage{subcaption}  
\usepackage[english]{babel}

\usepackage{amsmath}

\def\tsc#1{\csdef{#1}{\textsc{\lowercase{#1}}\xspace}}
\tsc{WGM}
\tsc{QE}
\tsc{EP}
\tsc{PMS}
\tsc{BEC}
\tsc{DE}

\begin{document}
\renewcommand{\floatpagefraction}{.7}
\renewcommand{\textfraction}{.1}

\shorttitle{Trustworthy and Explainable Deep Reinforcement Learning for Safe and Efficient Process Control}
\shortauthors{V. Bezold, A. Sauer}

\title [mode = title]{Trustworthy and Explainable Deep Reinforcement Learning for Safe and Energy-Efficient Process Control: A Use Case in Industrial Compressed Air Systems}                      

\affiliation[1]{organization={Fraunhofer Institute for Manufacturing Engineering and Automation IPA},
                addressline={Nobelstraße 12}, 
                city={Stuttgart},
                postcode={70569}, 
                country={Germany}}
\affiliation[2]{organization={Institute for Energy Efficiency in Production EEP, University of Stuttgart},
                addressline={Nobelstraße 12}, 
                city={Stuttgart},
                postcode={70569}, 
                country={Germany}}
                
\affiliation[3]{organization={Institute of Industrial Manufacturing and Management IFF, University of Stuttgart},
                addressline={Allmandring 35}, 
                city={Stuttgart},
                postcode={70569}, 
                country={Germany}}

\author[1,2]{Vincent Bezold}[type=editor,
                        auid=000,
                        bioid=1,
                        orcid=0009-0006-4430-4710]

\cormark[1]
\ead{vincent.bezold@ipa.fraunhofer.de}

\credit{Data Curation, Methodology, Software, Validation, Visualization, Writing – Original Draft Preparation, Writing – Review \& Editing}

\author[1]{Patrick Wagner}[type=editor,
                        auid=000,
                        bioid=2,
                        orcid=]

\ead{patrick.wagner@ipa.fraunhofer.de}

\credit{Methodology, Writing – Review \& Editing}

\author{Jakob Hofmann}[type=editor,
                        auid=000,
                        bioid=3,
                        orcid=0009-0004-0871-8444]

\credit{Data Curation, Methodology, Software, Validation}
\author[1,3]{Marco Huber}[type=editor,
                        auid=000,
                        bioid=4,
                        orcid=0000-0003-3413-6291]

\credit{Funding Acquisition, Project Administration, Resources, Writing – Review \& Editing}                
\author[1,2]{Alexander Sauer}[type=editor,
                        auid=000,
                        bioid=4,
                        orcid=0000-0003-3822-1514]

\credit{Funding Acquisition, Project Administration, Resources, Writing – Review \& Editing}
\cortext[cor1]{Corresponding author}



\definecolor{codegreen}{rgb}{0,0.6,0}
\definecolor{codegray}{rgb}{0.5,0.5,0.5}
\definecolor{codepurple}{rgb}{0.58,0,0.82}
\definecolor{backcolour}{rgb}{0.95,0.95,0.92}





\begin{abstract}
This paper presents a trustworthy reinforcement learning approach for the control of industrial compressed air systems. We develop a framework that enables safe and energy-efficient operation under realistic boundary conditions and introduce a multi-level explainability pipeline combining input perturbation tests, gradient-based sensitivity analysis, and SHAP (SHapley Additive exPlanations) feature attribution. An empirical evaluation across multiple compressor configurations shows that the learned policy is physically plausible, anticipates future demand, and consistently respects system boundaries. Compared to the installed industrial controller, the proposed approach reduces unnecessary overpressure and achieves energy savings of approximately 4\,\% without relying on explicit physics models. The results further indicate that system pressure and forecast information dominate policy decisions, while compressor-level inputs play a secondary role. Overall, the combination of efficiency gains, predictive behavior, and transparent validation supports the trustworthy deployment of reinforcement learning in industrial energy systems.
\end{abstract}

\begin{graphicalabstract}
\centering
\includegraphics[width=\textwidth,keepaspectratio]{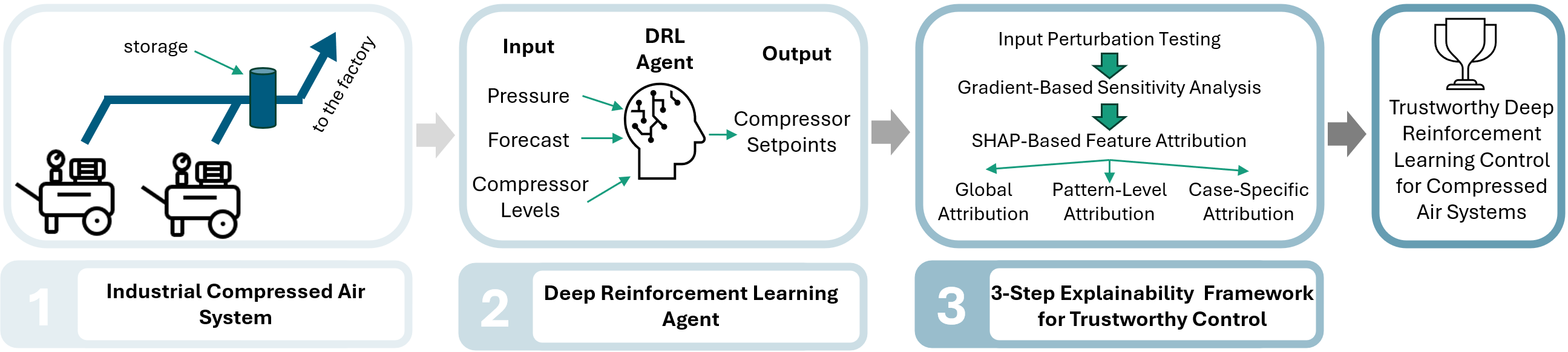}
\end{graphicalabstract}

\begin{highlights}
\item Interpretable deep RL for safe and efficient compressed air system control.
\item Multi-level explainability with SHAP, sensitivity, and perturbation tests.
\item Physically plausible and predictive policy respecting system boundaries.
\end{highlights}

\begin{keywords}
Deep Reinforcement Learning \sep Trustworthy Artificial Intelligence \sep Explainable Artificial Intelligence \sep Safe Reinforcement Learning \sep Industrial Process Control \sep Compressed Air Systems \sep Explainability \sep SHAP
\end{keywords}

\maketitle

\section{Introduction}

Compressed air systems are integral to many industrial processes and account for a significant share of electricity consumption in manufacturing environments \citep{DOE2017, Saidur2010}. Optimizing their operation is crucial for reducing energy demand, improving reliability, and enabling adaptive production planning \citep{DOE2017}.

Conventional control strategies, such as rule-based logic and model predictive control, often struggle with the nonlinear dynamics, component switching, and load variability inherent in multi-compressor systems \citep{Egan2023}. Reinforcement learning (RL) has emerged as a promising alternative, capable of learning adaptive control policies directly from interaction data \citep{Sutton2018}. It has shown success across diverse domains including robotics, games, recommendation systems, and language model fine-tuning through human feedback \citep{Christiano2017, Levine2023, Ouyang2022}.

However, the adoption of RL in industrial control remains limited. Key concerns include the black-box nature of deep RL agents, sensitivity to distributional shift, and a lack of verifiability \citep{Cheng2025, DulacArnold2021, Garcia2015}. Another fundamental challenge is the data-hungry and exploratory nature of RL, which makes learning directly on physical systems impractical or even infeasible due to safety and reliability concerns \citep{gu2024enhancingefficiencysafereinforcement}. As a result, simulation environments are typically required to enable agent training before deployment in real-world applications.

To address the issues of opacity and lack of trust in learned policies, the field of explainable artificial intelligence (XAI) has developed a variety of methods that enhance the transparency and interpretability of complex machine learning models, including RL agents. These techniques provide systematic ways to understand, validate, and build confidence in agent behavior, which is crucial for industrial deployment.

In this work, we investigate deterministic RL agents for compressor control with a focus on physical plausibility, interpretability, and policy transparency. We introduce a structured evaluation framework that combines input perturbation tests, gradient-based sensitivity analysis, and post hoc explainability using SHAP (SHapley Additive exPlanations) \citep{Beechey2023, Lundberg2017SHAP}. This enables us to systematically assess both the decision quality and the internal reasoning of the policy across a wide range of operating conditions \citep{McMahan2024}.

We evaluate our approach across multiple compressor setups and show that the RL agent learns to anticipate demand, operate within system constraints, and reduce energy use without relying on handcrafted control models \citep{Achiam2017, Cauz2024}. While these results support the viability of RL for industrial control, further research is needed to ensure robustness under rare and adversarial scenarios before deployment in safety-critical systems can be fully endorsed.

\section{Related Work}
\label{sec:relatedWork}

Reinforcement learning has become a core methodology in modern artificial intelligence, with applications ranging from robotics and game-playing to energy systems and natural language processing \citep{Sutton2018, Levine2023, Ouyang2022}. Deep reinforcement learning (DRL), in particular, enables scalable policy optimization in complex environments with high-dimensional inputs and nonlinear dynamics.

\textbf{Compressed Air and Industrial Systems:}  
Compressed air systems have recently received increased attention in the RL community due to their high energy demand and control complexity. \citet{zhong2024fuzzy} integrate interval type-2 fuzzy logic with DRL to improve robustness under uncertainty. \citet{daneshvar2023safe} combine imitation learning and primal-dual optimization for safe control of compressed air energy storage. \citet{Li2023HierarchicalRL} propose a hierarchical RL controller for multi-compressor coordination that explicitly accounts for component degradation. 

\textbf{Cross-Domain Applications:}  
In related domains, DRL has been applied to HVAC (Heating, Ventilation, and Air Conditioning) systems, microgrids, and manufacturing processes. For instance, \citet{azuatalam2020hvac} apply the Deep Deterministic Policy Gradient (DDPG) algorithm to HVAC demand response. \citet{delavari2024microgrid} use RL to coordinate hybrid microgrid storage. General industrial applications are covered by \citet{Gomes2024ReinforcementL} and \citet{Qiu2025ASO}. In the mobility domain, \citet{10186787} propose an RL-based energy management strategy for hybrid electric vehicles. Furthermore, large-scale deployments such as DeepMind’s use of RL to optimize data center cooling have demonstrated real-world impact and up to 40\% efficiency gains \citep{deepmind2022cooling}.

\textbf{Explainability and Trustworthiness:}  
As DRL policies are inherently complex and nonlinear, explainability becomes essential for trust and deployment. A variety of approaches have been proposed, including gradient-based analysis, surrogate modeling, and model-agnostic feature attribution. Among the latter, SHAP (SHapley Additive exPlanations) has gained popularity due to its theoretical grounding and flexibility \citep{Lundberg2017SHAP}. \citet{Pfeiffer2024DemystifyingRL} use DeepSHAP to interpret DRL scheduling agents. \citet{Suryavanshi2022ASO} survey post hoc and intrinsic explainability methods in RL, including SHAP and feature saliency maps. \citet{Brandes2024distill2explainDD} and \citet{haeussler2025xai} propose inherently interpretable surrogates such as differentiable decision trees and symbolic rule tables. While promising, these methods may only generalize well in dense regions of the training data, with reduced reliability under distributional shift.

\textbf{Safe and Predictive Control:}  
Several works incorporate safety into RL via control-theoretic constraints or hybrid architectures. \citet{cocaul2025safe} embed Control Lyapunov and Barrier Functions into Soft Actor-Critic (SAC) agents to ensure formal safety guarantees. \citet{bloor2024controlinformed} hybridize PID control with DRL for enhanced robustness. Model Predictive Control (MPC) has also been combined with DRL, as in \citet{Onyenanu2024IntegratingMP} and \citet{Mohamed2025LeveragingRL}, who use Koopman-based dynamics to improve DRL-enhanced MPC.

\textbf{Verification and Generalization:}  
Verification tools for DRL include reachability analysis, model checking, and adversarial testing. \citet{10.1145/3596444} provide a survey of such techniques. \citet{Mayer2025MultiOR} promote multi-objective optimization to balance efficiency and robustness. \citet{zhan2025datacentercoolingoptimization} apply offline RL for safe and efficient cooling control in data centers, using physics-informed constraints.

\textbf{Positioning:}  
These contributions demonstrate DRL’s increasing maturity across domains, including energy and industrial control. However, explainability and direct behavioral validation remain underexplored for compressed air systems. Our work addresses this gap through a structured evaluation framework combining SHAP-based attribution and deterministic input perturbation tests. This contributes toward a more transparent and trustworthy deployment of DRL in constrained, safety-relevant settings.

\section{Scope and Contributions}
\label{sec:scope_and_contributions}

This work addresses the problem of trustworthy control of compressed air systems using reinforcement learning. The focus lies on training policies that operate reliably under physical and technical constraints while maintaining interpretability and operational transparency. The study targets industrial multi-compressor setups with variable demand and safety-critical boundaries, where robust and explainable behavior is essential for real-world adoption.

The main contributions of this paper are as follows:
The main contributions of this paper are as follows:
\begin{itemize}
    \item A reinforcement learning framework for compressed air systems that ensures safe and efficient control under industrial operating constraints.
    \item A multi-level explainability pipeline combining SHAP, gradient-based sensitivity, and input perturbation tests to provide transparent and verifiable insights into policy behavior, including a time-resolved attribution aligned with representative control trajectories.
    \item An empirical evaluation across multiple system configurations showing that the learned policy is physically plausible, anticipates demand, respects system boundaries, and improves energetic efficiency without relying on handcrafted control models.
\end{itemize}

Together, these contributions form a foundation for interpretable and trustworthy reinforcement learning in industrial energy systems and provide a transferable methodology for future applications in adjacent domains.


\section{Methodology}
\label{sec:methodology}

The methodology comprises four sequential steps: (1) constructing the simulation environment; (2) developing and training the RL agent; (3) evaluating its control performance against a baseline; and (4) applying explainability analyses to the learned policy.
\subsection{Environment Construction and Experimental setup}
\label{sec:environment_setup}

This section describes the construction of the simulation environment used for training and evaluating the reinforcement learning agent. The environment integrates compressor-specific and system-level technical components, a physical model of the compressed air dynamics, and a reward function that encodes operational objectives and constraints. These elements together define the state and action spaces available to the agent and establish the closed-loop interaction cycle between agent and system.

\subsubsection{Technical Components}

The simulation environment is composed of compressor-specific and system-specific technical components. Compres\-sor-specific elements depend on the compressor type: for fixed-speed compressors, key parameters include the fixed volumetric flow rate at the operational point and the allowed number of switching cycles per hour, provided by the manufacturer. Variable-speed compressors require characterization by their power-to-flow curves, typically provided by manufacturers or derived from empirical data fitting, as was performed in this study. Specific parameter values are not disclosed here due to confidentiality.

System-level technical parameters include permissible operational pressure ranges, storage tank volumes, and associated system constraints. These parameters define boundary conditions and significantly impact the dynamic behavior modeled within the environment.

\subsubsection{System Modeling}

The compressed air system dynamics were modeled based on fundamental thermodynamic and physical laws, primarily leveraging the ideal gas equation to realistically capture system responses. This approach is justified because the operating pressures in industrial compressed air systems are typically only a few bars above atmospheric pressure, and the air temperature remains close to ambient. Under these conditions, deviations from ideal gas behavior are negligible, allowing the use of the ideal gas law without significant loss of accuracy \citep{Moran2014}. Central to the dynamic modeling is the storage tank pressure \( p(t) \) (a state space variable in the RL system), which evolves over discrete time steps according to the mass balance equation expressed as:

\begin{equation}
    p(t+\Delta t) = p(t) + \frac{\Delta V \cdot p(t)}{V_\text{storage}}~,
\end{equation}

\noindent where \( \Delta V \) represents the net volume change of air within the tank during timestep \( \Delta t \), and \( V_\text{storage} \) denotes the total storage volume of the system. The net volume flow is computed as the difference between the aggregated volumetric outputs from all active compressors and the scaled consumer demand. Compressor outputs differ based on type: fixed compressors operate in discrete states (on/off), while variable-speed compressors continuously adjust their outputs within predefined operational limits.

To facilitate learning, forecasted demand values are normalized before being passed to the agent, while all calculations within the environment, including rewards and pressure updates, are performed using the original (unnormalized) physical values.

\subsubsection{Reward Function Design}

The reward function guides the reinforcement learning agent toward minimizing energy consumption while enforcing operational constraints on system behavior. Energy-related operational costs \( C_{\text{energy}} \) constitute the primary component of the reward function, computed as:

\begin{equation}
    C_{\text{energy}} = \sum_{i=1}^{N} P_i \cdot \frac{\Delta t}{3600} \cdot c_{\text{electricity}}~,
\end{equation}

\noindent where \( P_i \) represents the electrical power consumption of compressor \( i \), \( \Delta t \) is the timestep duration in seconds, and \( c_{\text{electricity}} \) denotes the constant electricity price in €/kWh.

To enforce operational constraints, the following two penalty components are incorporated

\textbf{Pressure Penalty:}  
The pressure penalty \( P_{\text{pressure}} \) discourages deviations from the operational reference pressure \( p_\text{ref} \). Specifically, pressure deviations exceeding this reference are penalized proportionally, as defined by:

\begin{equation}
    P_{\text{pressure}} = \alpha_{\text{penalty}} \cdot \max\left(0, \frac{p_{\text{actual}}}{p_{\text{ref}}} - 1\right)~,
\end{equation}

\noindent where \( \alpha_{\text{penalty}} \) is a predefined penalty coefficient, \( p_{\text{actual}} \) is the actual tank pressure at the given timestep, and \( p_{\text{ref}} \) represents the desired operational reference pressure.

\textbf{Switching Penalty:}  
The switching penalty \( P_{\text{switching}} \) addresses operational constraints of fixed-speed compressors, which have manufacturer-specified limits on how frequently they can be turned on and off within an hour. This limit is enforced through a rolling switch-counter mechanism: each compressor has an hourly switching allowance, incrementally refilled each timestep. A penalty is applied if the compressor exceeds the allowed switching rate, discouraging frequent switching and encouraging smoother operational patterns.

The complete reward function thus becomes

\begin{equation}
    R = -(C_{\text{energy}} + P_{\text{pressure}} + P_{\text{switching}})~.
\end{equation}

\noindent This design steers the agent toward cost-effective, constraint-compliant, and persistent control strategies. 
To assess the robustness of the reward formulation, the weighting factors of the individual penalty terms are varied in preliminary experiments. The pressure and switching penalty coefficients are adjusted to explore different trade-offs between energy efficiency, pressure stability, and switching frequency. The final weights are selected based on these trials to balance efficiency and operational smoothness.

\subsubsection{Agent Development and Training}
\label{subsec:agent_training}

The development and training of the reinforcement learning agent comprise three main aspects: the selection of a suitable algorithm for continuous control, the design of the underlying neural network architecture and hyperparameters, and the definition of the training procedure. Each of these components is outlined in the following.

\subsubsection*{Algorithm Selection}  
For continuous control tasks, reinforcement learning algorithms such as Deep Deterministic Policy Gradient (DDPG), Soft Actor-Critic (SAC), and Proximal Policy Optimization (PPO) are commonly employed. The choice of algorithm is guided by three criteria: stability under non-linear dynamics, sample efficiency, and implementation maturity in state-of-the-art frameworks. Based on these criteria, two representative candidates (SAC and PPO) were shortlisted for further comparison, while DDPG was not considered due to limited robustness and support in modern toolkits. The comparative evaluation of SAC and PPO is presented in Section~\ref{subsec:agent_training_results}.

\subsubsection*{Network Architecture and Hyperparameters}  
The policy and value functions are represented by fully connected neural networks augmented with a recurrent Long Short-Term Memory (LSTM) component to capture temporal dependencies. The number of layers, neurons per layer, and activation functions were chosen to provide sufficient expressive power while avoiding excessive model complexity. Hyperparameters such as learning rate, discount factor, and clipping range were tuned in preliminary trials to ensure stable training behavior. The final architecture and parameter values are summarized in Table~\ref{tab:hyperparameters}.

\subsubsection*{Training Setup}  
The training procedure relies on parallelized environment rollouts to efficiently generate diverse state transitions. Gradient updates are performed on mini-batches sampled from these rollouts using GPU acceleration. Termination criteria include a fixed maximum number of training iterations as well as early stopping if performance stabilizes. This setup ensures reproducibility and prevents overfitting while keeping computational effort manageable.
\subsection{Agent Results and Evaluation}
\label{subsec:Agentevaluation}

The trained agent is evaluated against the baseline given by the industrial controller installed in the real system. Performance is assessed using metrics such as cumulative reward, deviation from reference setpoints, and frequency of constraint violations. This evaluation highlights the extent to which the learned policy improves operational efficiency and respects system constraints, while also revealing potential shortcomings or edge cases for further analysis.

\subsection{Explainability Framework}
\label{subsec:explainability}
To interpret the learned policy $\pi_\theta$, we design a three-step explainability pipeline that combines direct policy probing, gradient-based sensitivity analysis, and SHAP-based feature attribution. These techniques jointly capture functional responses, directional sensitivities, and feature-level relevance.

To maintain readability, mathematical derivations for the sensitivity gradients and SHAP values are provided in Appendix~\ref{app:explainability_math}.

\subsubsection*{Input Perturbation Testing}
This approach, sometimes called scenario-based policy probing or counterfactual analysis, probes the policy's behavior under carefully designed, static state configurations rather than using a standard test dataset. Each configuration is constructed to reflect meaningful physical scenarios (e.g., varying pressure levels with increasing forecasted flow). The goal is not statistical generalization, but deterministic inspection of plausibility and control logic. 

This differs from test set evaluation in that:
\begin{itemize}
    \item It uses hand-crafted, often extreme or edge-case inputs not present in training data.
    \item It focuses on \emph{functional response} rather than predictive accuracy.
    \item It enables visualization of policy smoothness and discrete activation thresholds (e.g., compressor switching).
\end{itemize}

\subsubsection*{Gradient-Based Sensitivity Analysis}
To assess the importance of each input feature for the agent's decisions, we employ gradient-based sensitivity analysis. In this context, the term \textit{saliency} refers to the local sensitivity of the policy output to small changes in each input feature, quantified by the gradient (i.e., the partial derivative) of the policy with respect to that feature. We compute these gradients using automatic differentiation. This provides a directional measure of local sensitivity, how much a small change in an input alters the action. For a policy $\pi_\theta(s)$ and input state $s \in \mathbb{R}^{n_s}$, the gradient magnitude is:

\[
g_j(s) = \left|\frac{\partial \pi_\theta(s)}{\partial s_j}\right|
\quad \text{for each feature } j~.
\]

Averaging over $N$ sampled states yields a saliency value (or relevance score) per feature:

\[
G_j = \frac{1}{N} \sum_{i=1}^{N} g_j(s_i)~.
\]

\noindent
This analysis reveals which features dominate the policy decision function, typically pressure and short-term forecast, and whether feature importance shifts with increasing state complexity.

\subsubsection*{SHAP-Based Feature Attribution}
To complement gradients with attribution scores ground-ed in cooperative game theory, we apply SHAP (SHapley Additive exPlanations). SHAP decomposes the output into additive contributions from each input feature and is suitable for interpreting black-box models like neural policies.

We structure our SHAP analysis across three resolution levels:

\begin{itemize}
    \item \textbf{Global Attribution:} Aggregated SHAP values over many inputs provide average feature relevance. This helps identify consistently important inputs such as pressure or forecast.
    
    \item \textbf{Pattern-Level Attribution:} Scatter plots visualize how SHAP values vary with raw feature values. This reveals functional trends (e.g., linear, saturating, non-monotonic), helping assess whether the agent has learned physically plausible mappings.
    
    \item \textbf{Case-Specific Attribution:} Waterfall plots explain single decisions by showing how individual input values contributed to the total output. This level is particularly useful for validating edge-case decisions and debugging.
\end{itemize}
\subsubsection*{SHAP-Based Feature Attribution}
To complement gradients with attribution scores ground-ed in cooperative game theory, we apply SHAP (SHapley Additive exPlanations). SHAP decomposes the output into additive contributions from each input feature and is suitable for interpreting black-box models like neural policies.

We structure our SHAP analysis across three resolution levels:

\begin{itemize}
    \item \textbf{Global Attribution:} Aggregated SHAP values over many inputs provide average feature relevance. This helps identify consistently important inputs such as pressure or forecast.
    
    \item \textbf{Pattern-Level Attribution:} Scatter plots visualize how SHAP values vary with raw feature values. This reveals functional trends (e.g., linear, saturating, non-monotonic), helping assess whether the agent has learned physically plausible mappings.
    
    \item \textbf{Case-Specific Attribution:} Waterfall plots explain single decisions by showing how individual input values contributed to the total output. This level is particularly useful for validating edge-case decisions and debugging.
\end{itemize}

\textbf{Time-Resolved SHAP Attribution:}
To complement static SHAP analyses, we additionally perform a time-resolved attribution aligned with representative control schedules. Although the policy is Markovian, the resulting closed-loop behavior emerges dynamically through repeated interaction with the system dynamics. To capture this effect, SHAP values are computed at each time step for synthetically designed pressure and demand sweep scenarios and visualized alongside the corresponding excitation signal. This establishes a direct temporal link between systematic input variations and the evolving feature attributions. The resulting SHAP trajectories reveal how individual feature relevance changes over the course of a control sequence, enabling interpretation of dynamic decision principles and bridging static explainability with closed-loop control behavior.

\section{Experimental Setup}
\label{sec:experimental_setup}

The experimental setup comprises two variable-speed compressors, each rated at 30\,kW, and one fixed-speed compressor, also rated at 30\,kW. Due to confidentiality agreements, specific compressor models are not disclosed. The system includes a total compressed-air storage capacity of approximately \(5\,\text{m}^3\), including tanks and pipe volumes, servicing a tool manufacturing facility.

Operational data were collected over approximately two weeks at a sampling interval of 5 seconds. Measurements captured include system pressure, air temperature, total volumetric flow, individual compressor volumetric flows, and electrical power consumption per compressor. 

The configurable nature of the simulation environment enables extensive variation, such as changing the number of compressors or the length of the forecast horizon (i.e., the number of future consumption steps provided to the agent). For this study, configurations ranged from a minimal setup with one compressor and a single-step forecast to the realistic three-compressor system with forecast horizons of up to 5 steps.

\vspace{0.5em}
\noindent
To facilitate consistent reference throughout the paper, we introduce the following naming convention for experimental configurations. A configuration denoted as \textbf{XCYF} refers to a setup with \( X \) active compressors and a forecast horizon of \( Y \) timesteps. For example, \textbf{1C1F} corresponds to one compressor and a single forecast timestep, while \textbf{3C3F} describes three compressors and a forecast vector containing three future consumption values. 

In terms of compressor types, the 1C1F configuration uses a single variable-speed compressor. All 3CYF configurations, where \( X = 3 \) and \( Y \in \{1, 3, 5\} \), consist of one fixed-speed and two variable-speed compressors. This compact notation enables efficient referencing across experimental results and interpretability evaluations.

\section{Agent Training}
\label{subsec:agent_training_results}

\begin{figure}[h]
    \centering
    \includegraphics[width=1.0\linewidth]{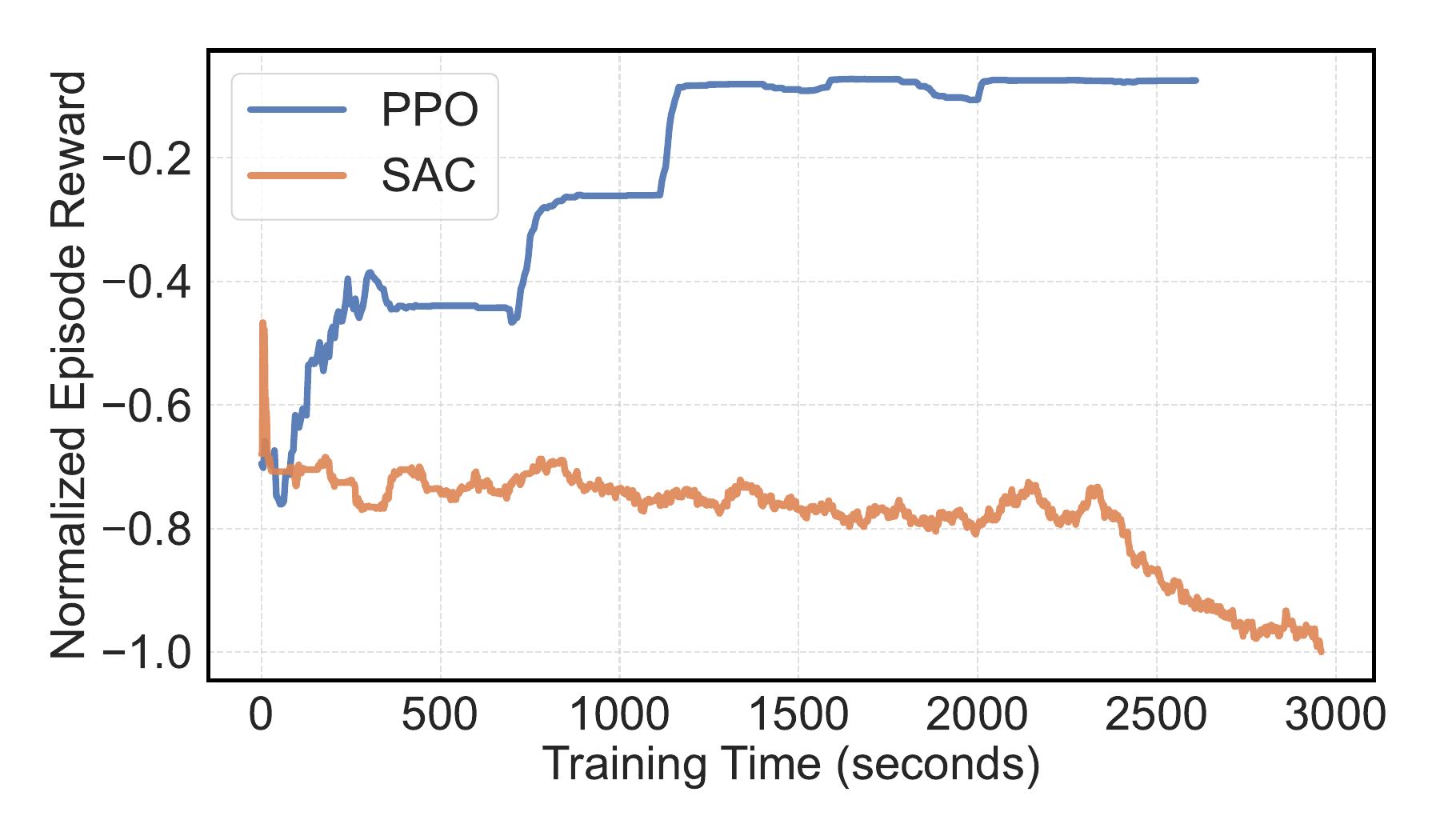}
    \caption{
        Comparison of convergence speed and stability for Proximal Policy Optimization (PPO) and Soft Actor-Critic (SAC) in the compressor control environment. The PPO algorithm exhibits substantially faster and more stable convergence, while SAC fails to reach comparable performance even after hyperparameter tuning.
    }
    \label{fig:ppo_vs_sac}
\end{figure}

Since the requirement for the algorithm is to handle continuous action spaces, this narrows down the selection to methods such as DDPG, SAC, and PPO, which are specifically designed for such problems~\citep{haarnoja2018soft, lillicrap2015continuous, schulman2017proximal}. However, in current practice, many earlier reinforcement learning algorithms are deprecated or no longer maintained in state-of-the-art frameworks. For example, RLlib, a widely used library for scalable RL, now primarily supports only PPO and SAC for continuous control tasks, while other algorithms like DDPG are deprecated or not actively recommended~\citep{sutton2018reinforcement, dulacarnold2021challenges, rllib_algos_2024}. Therefore, only the two most established algorithms, SAC and PPO, were compared in a quick head-to-head convergence test. The results are shown in Figure~\ref{fig:ppo_vs_sac}. As can be seen, PPO converges much faster, while SAC fails to converge reliably, even after some hyperparameter tuning. Therefore, all subsequent analyses in this work are based solely on PPO. The observed lack of convergence of SAC can be attributed to characteristics of the compressor control environment. The task involves hybrid dynamics with discrete switching behavior, non-linear system responses, and penalty-driven constraints, which are known to increase training variance for entropy-regularized off-policy methods. In contrast, PPO’s clipped policy updates and on-policy optimization provide more stable learning behavior in such constrained control settings. Since the focus of this work is not a comparative algorithmic study, SAC was not further pursued.

\subsection{Network Architecture and Hyperparameters}

The RL agent utilizes a neural network policy based on a fully connected architecture complemented by a Long Short-Term Memory (LSTM) unit, allowing the agent to capture temporal dependencies and dynamic system behavior effectively \cite{hochreiter1997lstm}. Due to the extensive parameter space and computational constraints, exhaustive hyperparameter optimization was not feasible. Therefore, guided heuristic trials informed by domain knowledge were performed to identify effective hyperparameters. The final selected hyperparameters, which yielded the best empirical performance, are summarized in Table~\ref{tab:hyperparameters} in the appendix.

\section{Results}
\label{sec:results}
This section presents the evaluation of the trained agent, followed by explainability analyses using input perturbation, gradient-based sensitivity, and SHAP feature attribution.

\subsection{Evaluation of Agent}
\label{subsec:evaluation}

The agent was trained entirely within the simulation environment introduced in Section~\ref{sec:environment_setup}, which was parameterized using measurements from the real factory system. To assess real-world applicability, the trained policy was evaluated on the original operational dataset, which captures actual consumption patterns and system responses under typical factory conditions. 

For a fair and meaningful assessment, the agent’s performance was compared to the control strategy that was in place during data collection. While the exact implementation details of this baseline controller are proprietary, it represents the installed industrial control logic of the system, which serves as the relevant real-world reference in this environment.

\begin{figure*}[htbp]
    \centering
    \includegraphics[width=1.0\linewidth]{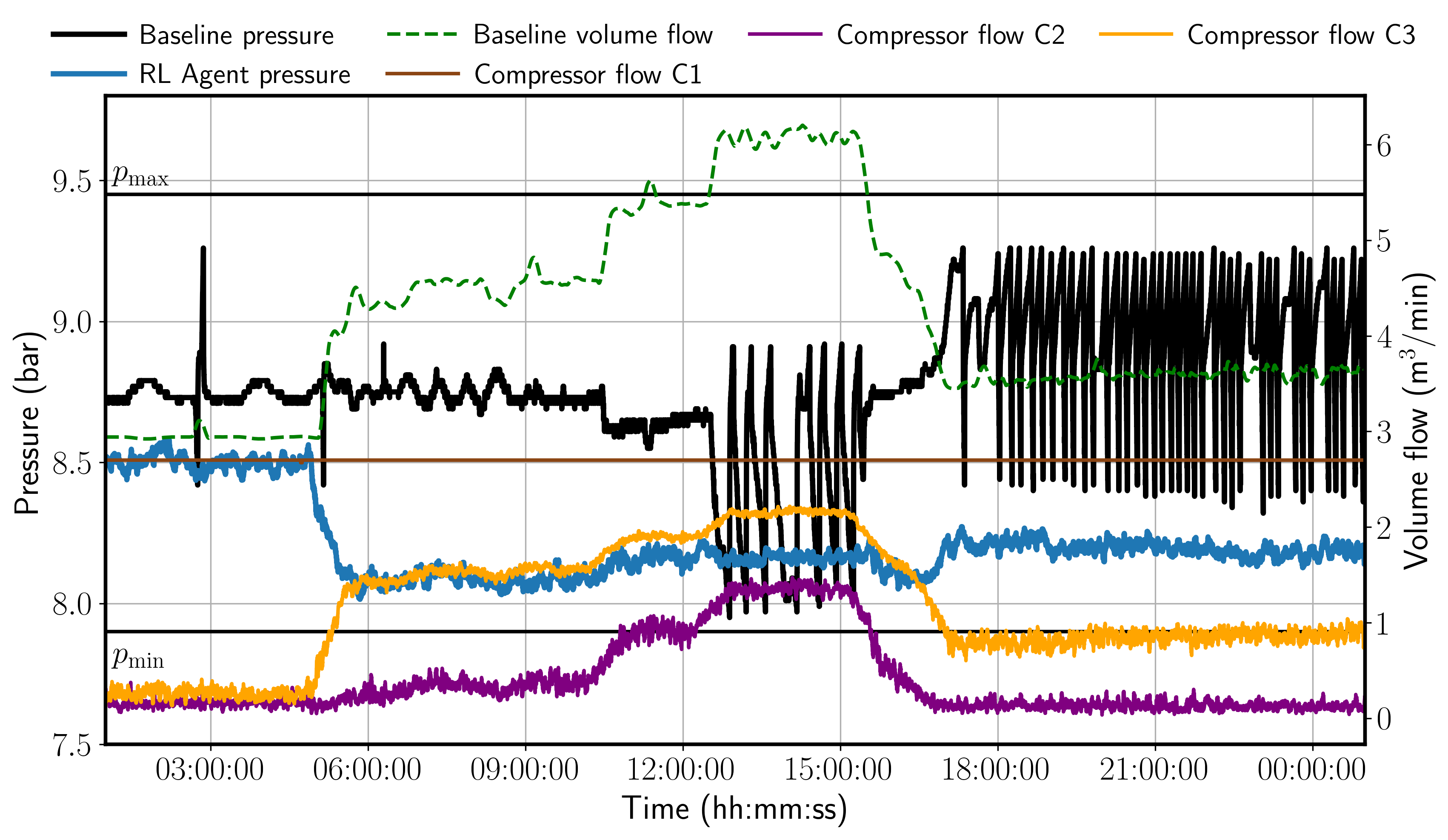}
    \caption{Comparison of system pressure and demand for a representative day based on real factory data, including the compressor-level supplied volumetric flow rates (control schedule). The RL-based optimized control maintains a lower average pressure than the installed industrial controller, while consistently adhering to the 8\,bar system limit. The additional compressor-level traces reveal how the RL policy allocates supply across individual compressors over time and avoids unnecessary overproduction that would increase pressure. This reduction in average pressure results in about 4\% energy savings.}
    \label{fig:pressure_vs_flow}
\end{figure*}

Figure~\ref{fig:pressure_vs_flow} visualizes a representative day of operation, displaying the actual volume flow demand, the system pressure maintained by the baseline control, and the pressure trajectory resulting from the RL-based optimized control, together with the resulting compressor-level supply schedule.

As shown, the RL agent consistently regulates system pressure at a lower average level than the baseline controller, while still reliably respecting the technical upper limit of 8\,bar. By reducing unnecessary overpressure, the RL-based control improves energetic efficiency without sacrificing operational safety. In this scenario, the RL agent achieves an energy savings of approximately 4\% relative to the existing baseline, demonstrating the practical benefits of data-driven optimization in a real industrial context.
Additional experiments with varied reward weights indicate that the learned policy is robust to moderate changes in the individual penalty terms. Different weightings mainly affect the trade-off between energy consumption, pressure regulation, and switching frequency, while the qualitative control strategy and anticipatory behavior remain unchanged.

Beyond aggregate pressure and demand trajectories, the compressor-level volumetric flow signals provide a detailed \emph{control schedule} that makes the operational differences between the baseline controller and the RL policy visible over time, including compressor activation patterns and load allocation across units.

While these initial results demonstrate the practical benefits of the RL-based control, a comprehensive performance evaluation across multiple days and operational scenarios is beyond the scope of this work. The focus of this paper is on the explainability and interpretability of the learned policy, rather than an extensive benchmarking of control performance. Further analyses of long-term performance and generalizability will be explored in future research.

\subsection{Policy Explainability}
\label{subsec:PolicyExplainability}

The following analyses investigate the explainability of the trained policy across different configurations. The three complementary approaches applied are input perturbation testing, gradient-based sensitivity analysis, and SHAP-based feature attribution. Together, these methods allow for both global and local interpretation of policy behavior and support the assessment of its physical plausibility.

\subsubsection{Input Perturbation Testing}
To evaluate the learned policy \(\pi_{\theta}\) in a deterministic and interpretable setting, an input perturbation testing procedure is applied. This method probes the agent’s control output across predefined input states, independently of environment feedback or stochastic effects. It serves as a sanity check to assess whether the control behavior follows physically meaningful patterns and responds intuitively to varying system conditions.

In this analysis, the policy output \( a = \pi_{\theta}(s) \) is evaluated over a sweep of forecasted consumer volumetric flow rates \(\dot{V}_\text{V}\), while keeping the system pressure \(p\) fixed at predefined values. The forecast sweep is sampled uniformly from the interval

\begin{equation}
\dot{V}_\text{V} \in \left[0, 3 \cdot \max_i(\dot{V}_{\text{max},i}) \right],
\end{equation}

\noindent
\noindent
where \(\dot{V}_{\text{max},i}\) denotes the maximum flow capacity of compressor \(i\). This covers the full spectrum from idle operation to strongly over-demanded scenarios. The fixed pressure values are selected from the set \(p \in \{7.9, 8.0, 8.1, 8.2, 8.3\}\,\text{bar}\). For each pair \((p, \dot{V}_\text{V})\), a state vector 
\(s = (p, \dot{V}_\text{V}, \text{levels})\) 
is constructed, where \(\text{levels}\) represents the current operating states of the compressors (normalized load levels for variable-speed machines and on/off status for fixed-speed machines). This state vector is then fed into the agent policy to determine compressor setpoints.

\begin{figure*}[h]
    \centering

    \begin{subfigure}[b]{0.48\textwidth}
        \includegraphics[width=\textwidth]{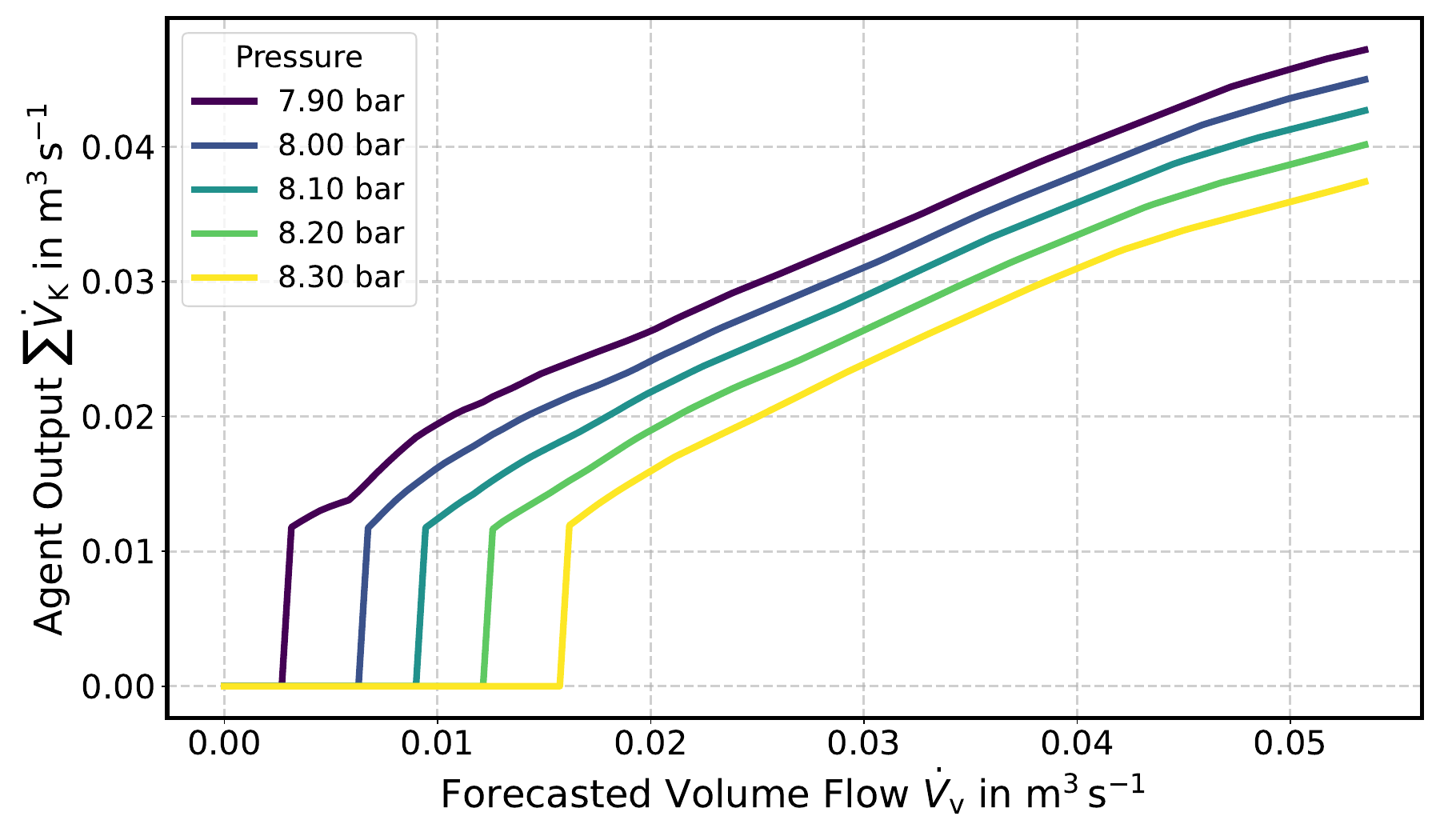}
        \caption{1C1F: Single variable-speed compressor with one-step forecast ($\SI{5}{s}$).}
    \end{subfigure}
    \hfill
    \begin{subfigure}[b]{0.48\textwidth}
        \includegraphics[width=\textwidth]{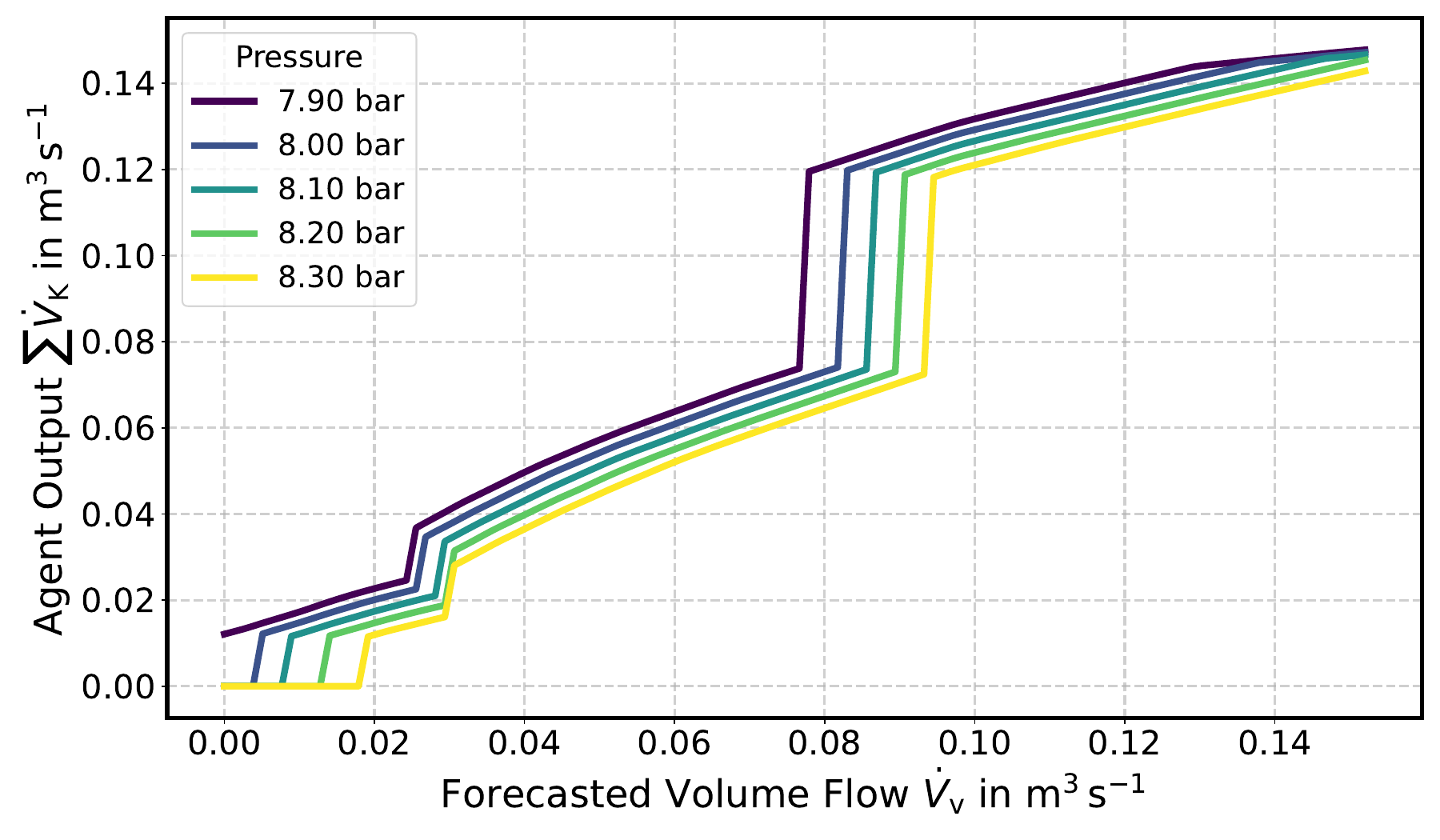}
        \caption{3C1F: One fixed-speed and two variable-speed compressors with one-step forecast ($\SI{5}{s}$).}
    \end{subfigure}

    \vspace{0.5cm}

    \begin{subfigure}[b]{0.48\textwidth}
        \includegraphics[width=\textwidth]{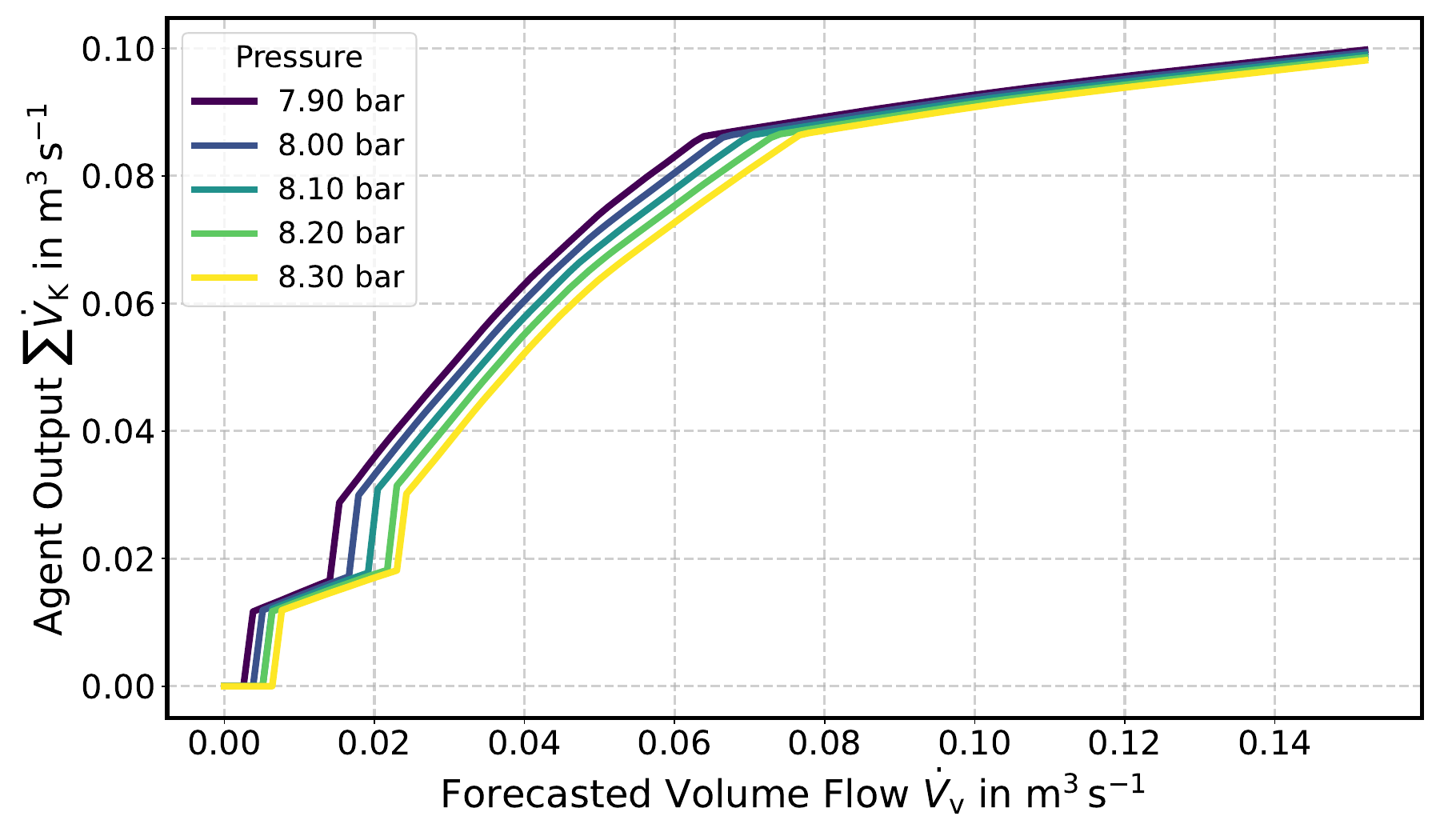}
        \caption{3C3F: Same compressor setup with a three-step forecast horizon ($\SI{15}{s}$).}
    \end{subfigure}
    \hfill
    \begin{subfigure}[b]{0.48\textwidth}
        \includegraphics[width=\textwidth]{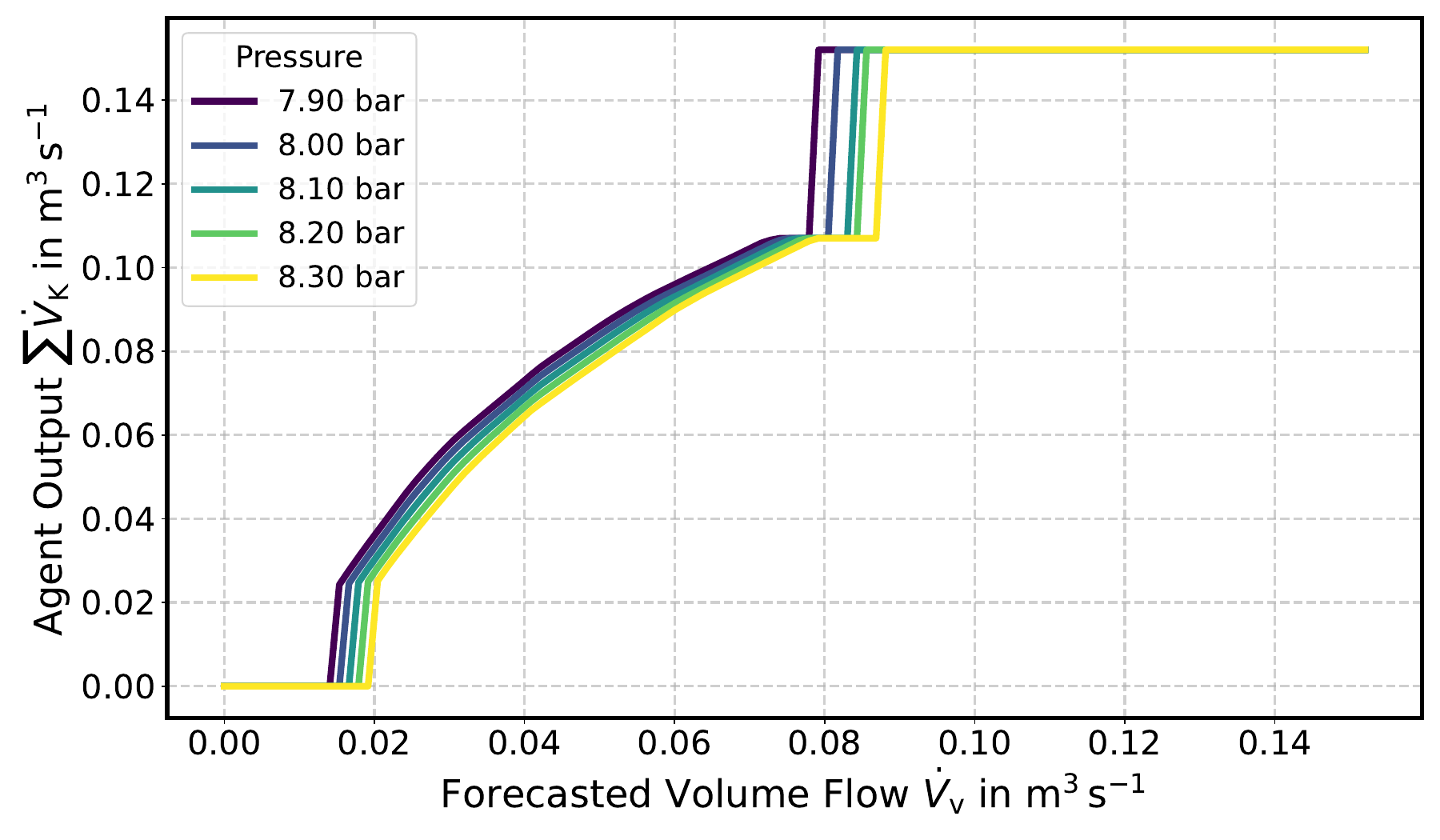}
        \caption{3C5F: Same compressor setup with a five-step forecast horizon ($\SI{25}{s}$).}
    \end{subfigure}

    \caption{Input perturbation testing across different configurations. Each curve shows the compressor output level selected by the agent in response to varying forecasted volumetric flow rates, under fixed pressure conditions.}
    \label{fig:direct_output}
\end{figure*}

The resulting policy outputs are visualized in Figure~\ref{fig:direct_output}. These curves reveal how the agent adjusts its control decisions in response to forecasted demand under fixed pressure conditions.

\textit{Interpretation}

The results confirm that the agent has learned a physically plausible control strategy. At lower pressure levels, it responds to rising flow demands with increased compressor output to stabilize the system. This trend appears consistently across all tested configurations.

In the simple 1C1F setting, the control response is smooth and monotonic, reflecting a direct mapping between predicted demand and compressor activity. In multi-compressor settings, the response curves exhibit discrete jumps. These stepwise changes indicate the agent’s internal logic for activating additional compressor units as demand increases. This behavior closely resembles real-world compressor scheduling, where machines are engaged incrementally once specific thresholds are crossed.

Overall, the input perturbation testing shows that the policy acts sensibly and predictably, validating the learned control behavior and providing additional interpretability beyond reward signals alone.

\subsubsection{Gradient-Based Sensitivity Analysis}

To complement the input perturbation testing, we apply gradient-based sensitivity analysis as described in Section~\ref{subsec:explainability}. This technique is used to identify which input features the agent’s policy \(\pi_{\theta}(s)\) is most sensitive to. Rather than analyzing individual decisions, this approach aggregates gradients over a large number of sampled states, enabling statistical insight into the overall feature relevance.

\textit{Experimental setup:}  
We generate \(N = 800\) random input states \(s_i\), each composed of three segments: (1) the normalized pressure, (2) a forecast vector of upcoming volumetric demand, and (3) the current levels of all compressors. The policy network's deterministic output is backpropagated with respect to each input using PyTorch’s automatic differentiation engine. The mean absolute gradients per input feature are then aggregated to form a sensitivity profile.

\begin{figure*}[h]
    \centering

    \begin{subfigure}[b]{0.48\textwidth}
        \includegraphics[width=\textwidth]{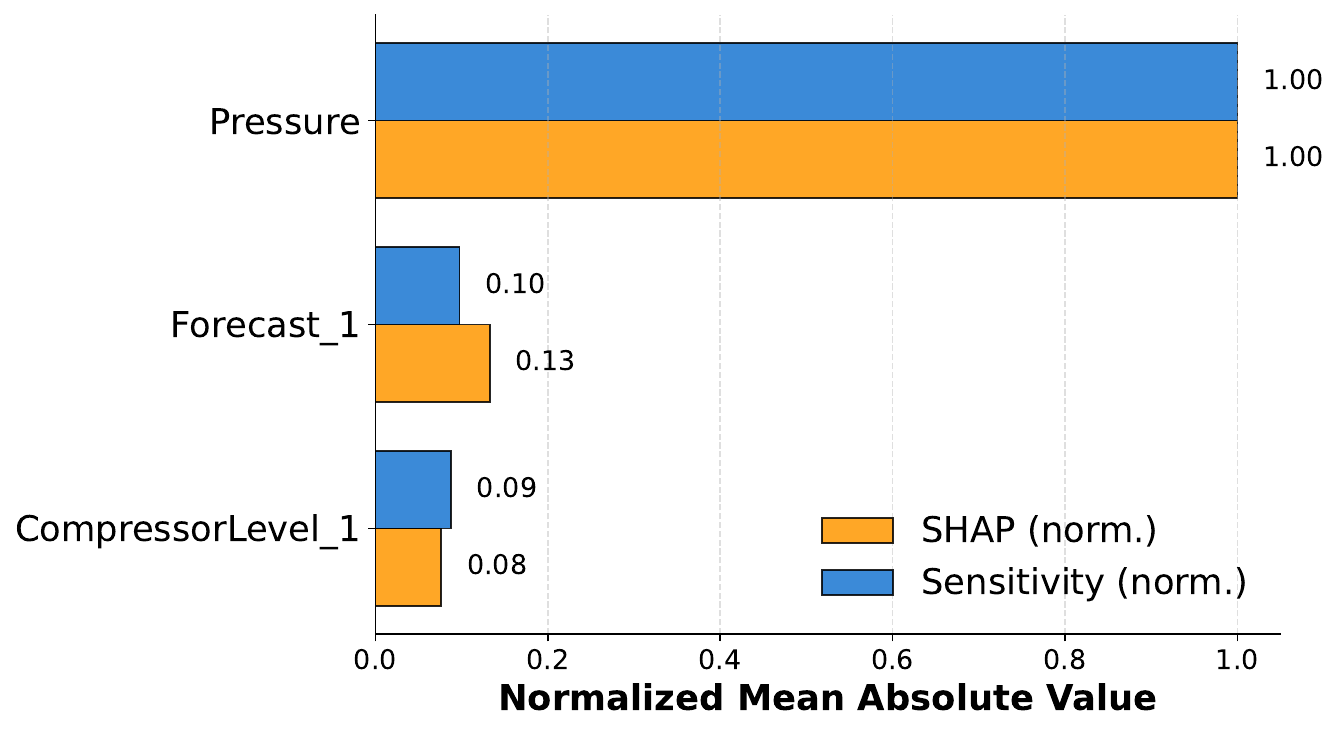}
        \caption{1C1F: Single variable-speed compressor, one-step forecast.}
    \end{subfigure}
    \hfill
    \begin{subfigure}[b]{0.48\textwidth}
        \includegraphics[width=\textwidth]{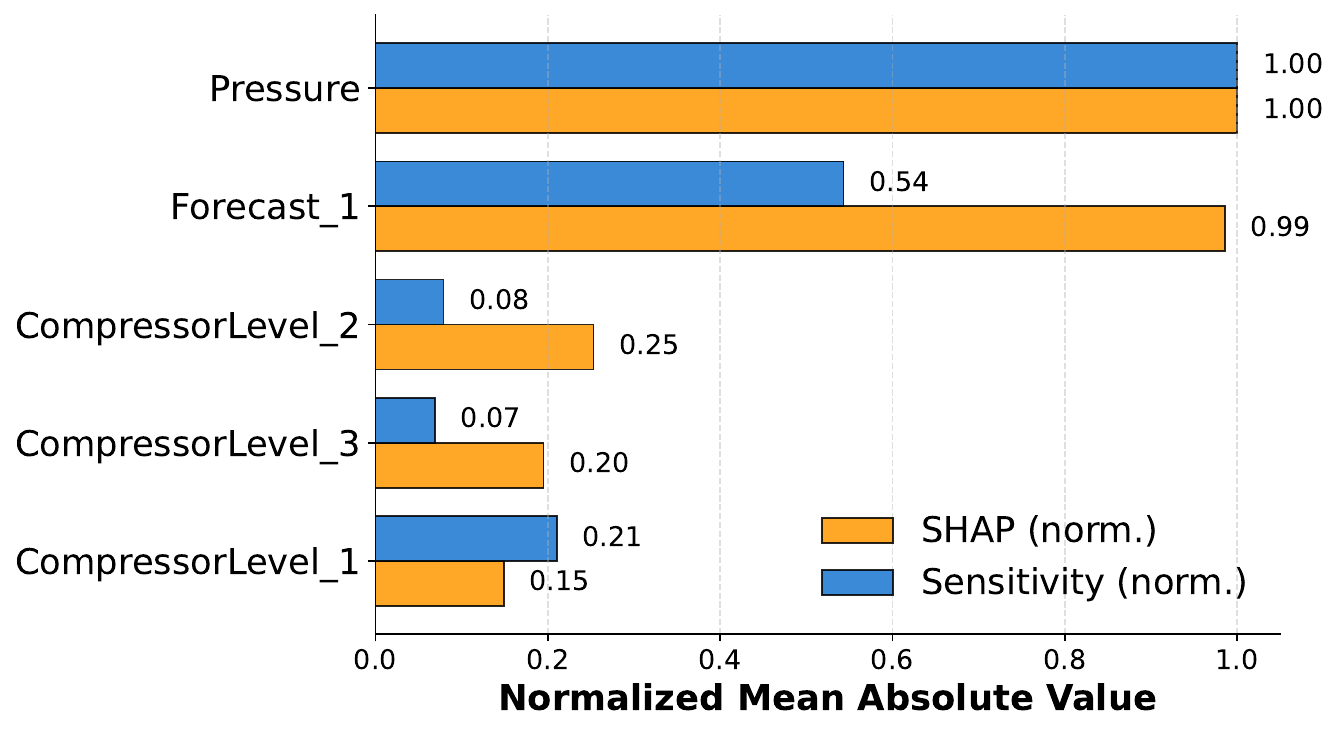}
        \caption{3C1F: One fixed-speed and two variable-speed compressors, one-step forecast.}
    \end{subfigure}

    \vspace{0.5cm}

    \begin{subfigure}[b]{0.48\textwidth}
        \includegraphics[width=\textwidth]{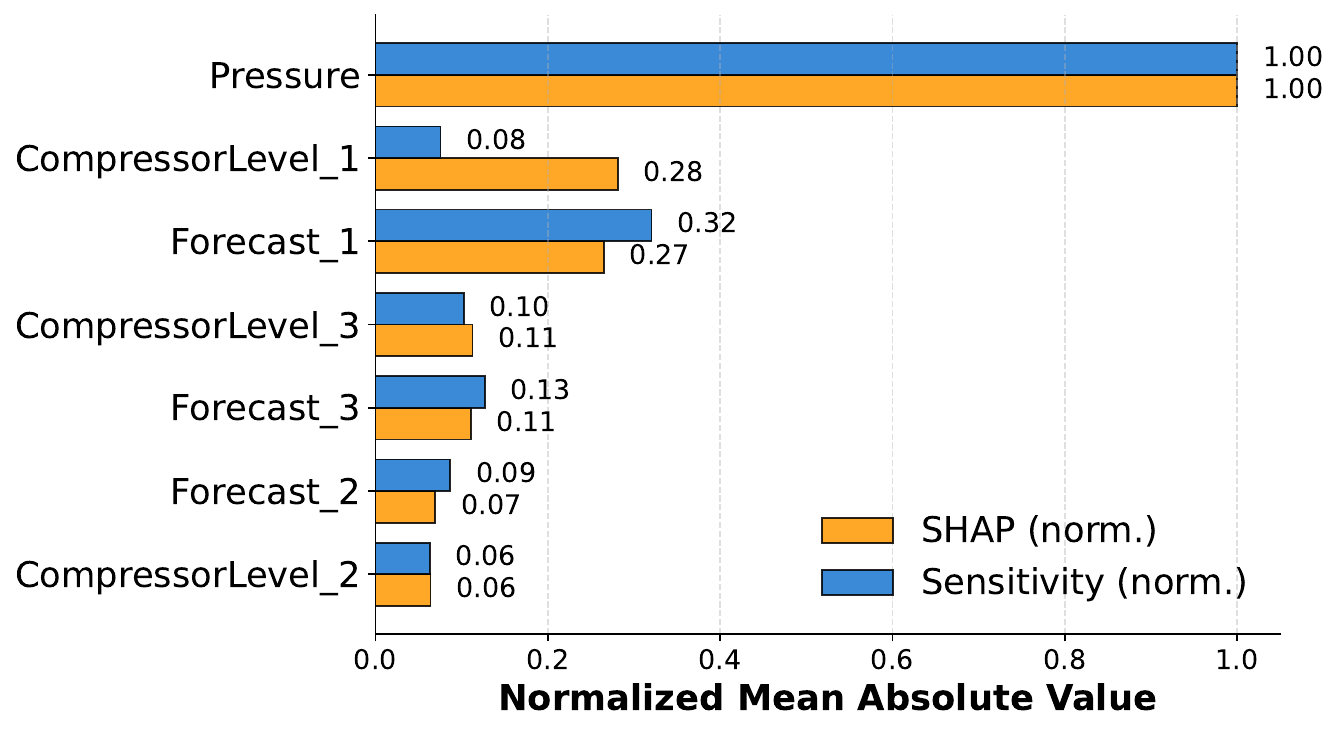}
        \caption{3C3F: Same compressor setup, three-step forecast.}
    \end{subfigure}
    \hfill
    \begin{subfigure}[b]{0.48\textwidth}
        \includegraphics[width=\textwidth]{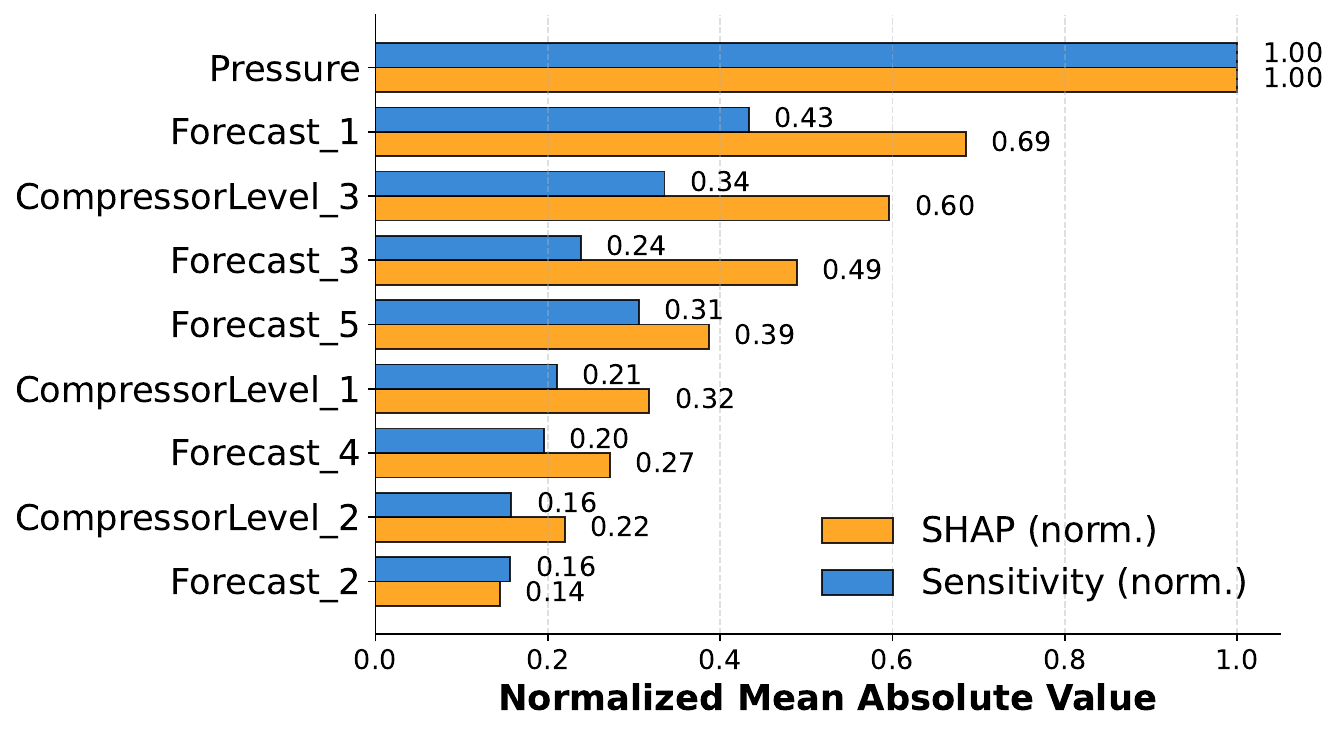}
        \caption{3C5F: Same compressor setup, five-step forecast.}
    \end{subfigure}

    \caption{
        Direct comparison of normalized mean absolute SHAP values (feature attribution) and gradient-based sensitivity scores (saliency) across all experimental configurations. Each bar plot visualizes both methods for each feature, allowing assessment of consistency in feature importance ranking. System pressure and the first forecasted demand consistently dominate in both approaches, supporting the physical plausibility and interpretability of the learned policy.
    }
    \label{fig:ShapVsSensitivity}
\end{figure*}

\textit{Results and interpretation:}  
The aggregated sensitivity profiles are presented in Figure~\ref{fig:ShapVsSensitivity} for all experimental configurations. Across all scenarios, regardless of compressor count or forecast horizon length, the system pressure consistently emerges as the most influential input feature. This finding aligns with the physical nature of compressed air systems, where pressure reflects the current balance between supply and demand. Since air flow responds directly to pressure gradients, this variable serves as a reliable and immediate indicator for control decisions.

In the minimal configuration 1C1F, where a single variable-speed compressor is controlled based on only one forecasted consumption value, the agent’s policy is almost exclusively sensitive to the current pressure. The forecast input and the control level both contribute minimally, with the forecast value being slightly more relevant than the control level. This reflects a direct and reactive control strategy, where the agent primarily tracks pressure to determine appropriate setpoints.

In the more complex 3CYF configurations (e.g., 3C1F, 3C3F, 3C5F), which involve one fixed-speed and two vari\-able-speed compressors, sensitivity to the first forecast value increases slightly but remains subordinate to the pressure signal. Additional forecast values beyond the first show only less contributions. Across all cases, compressor level inputs have consistently low sensitivity. This is a consequence of the control architecture: the agent outputs setpoints, which are interpreted and executed by subordinate controllers. As a result, the agent does not directly rely on the current actuator state but instead optimizes higher-level decisions.

Overall, the gradient-based sensitivity analysis confirms that the agent’s policy relies primarily on physically ground\-ed and temporally informative features—most notably, pressure and the immediate forecast—while internal system states such as current compressor levels play a minimal role. This supports the plausibility, robustness, and interpretability of the learned control strategy.
\newline
\newline
\textit{Interpretation of Saliency Values} \newline \indent
Gradient-based saliency scores reflect local sensitivity of the policy output to each input feature. Specifically, they quantify how strongly a small change in a feature affects the agent’s action output at a given state. For example, a high saliency value for the pressure input indicates that minor changes in pressure lead to large changes in the predicted compressor setpoints.

However, these values are not directly interpretable in terms of physical units or compressor activation levels. They express relative importance rather than absolute contribution. As such, saliency analysis is best suited for ranking features by influence and identifying directional dependencies in the policy function, rather than for explaining individual decisions in detail.

\subsubsection{SHAP Feature Attribution}

To further interpret the agent’s policy behavior, SHAP values are employed. SHAP provides post hoc, model-agnostic explanations by attributing the output of a deterministic model to its input features in a locally additive manner. In this context, the SHAP values represent contributions to the total action output of the policy, that is, the sum over all action dimensions at a given time step.

The analysis is structured into three levels of granularity as introduced in Section~\ref{subsec:explainability}: global importance across the entire dataset, local patterns revealed by SHAP scatter plots, and case-specific interpretability using SHAP waterfall plots for selected corner cases.
\newline
\newline
\textit{Experimental Setup} \newline \indent
Two sets of input states are generated to compute SHAP values. A background dataset with \(n_{\text{bg}} = 1024\) randomly sampled states is used to define a reference distribution. Additionally, a test set with \(n_{\text{test}} = 120\) states is sampled to evaluate SHAP values. Each state contains the normalized pressure, the forecast vector, and the current compressor levels. SHAP values are calculated using the deterministic agent policy as the model function. The SHAP attribution for each state reflects the contribution of each input feature to the summed policy output. For global and local analyses, SHAP values are aggregated across all test states.

\subsubsection*{Global Attribution}

The summary bar plots in Figure~\ref{fig:ShapVsSensitivity} display the mean absolute SHAP value per input feature. This reflects the overall importance of each feature in shaping the agent’s output. The pressure consistently emerges as the dominant feature, followed by the first forecast value. This finding holds across all tested configurations and reinforces earlier results obtained from gradient-based sensitivity analysis.

In more complex settings, such as the 3C3F and 3C5F configurations, the compressor levels begin to overlap in relevance with some forecast features. This is likely due to interaction effects in systems with higher control complexity. Nevertheless, the overall influence of the compressor level features remains modest when compared to the dominant input signals from pressure and forecast information.
\newline
\newline
\textit{Interpretation of SHAP Values} \newline \indent
SHAP values quantify the contribution of each input feature to the predicted agent output, relative to a reference (baseline) input. In the current analysis, both input features and outputs are normalized, so a SHAP value of 0.374 for the normalized pressure in the 1C1F configuration means that this feature increases the agent’s normalized control output by 0.374 units compared to the baseline. Since the output represents compressor setpoints normalized to their maximum allowable value, this indicates a substantial and direct influence on compressor activation.

A key strength of SHAP is its ability to provide additive, comparable explanations across features, regardless of their scale or units. High-magnitude SHAP values correspond to strong influence on the agent’s output in a given decision. While individual SHAP values can vary from one decision step to another, the summary bar plots in Figure~\ref{fig:ShapVsSensitivity} aggregate these values across many states and thus, reflect average feature importance within the agent’s policy.

Because normalization has been applied, the absolute values shown are best interpreted as relative contributions within the bounded (normalized) input and output spaces. This enables direct comparison of feature influence, but care should be taken when extrapolating these values to absolute system behavior.
\newline
\newline
\textit{Comparison of SHAP and Saliency Values} \newline \indent
The direct comparison in Figure~\ref{fig:ShapVsSensitivity} demonstrates that both SHAP and gradient-based sensitivity analyses provide a largely consistent ranking of feature importance. Across most scenarios, system pressure and the first forecasted demand value emerge as the most influential input features, while compressor-level features generally exhibit lower importance. However, in the 3C3F configuration, SHAP attribution indicates that compressor level 1 contributes more on average than the first forecast value. In contrast, the sensitivity analysis for the same scenario still ranks the forecast value above all compressor-level features. 

Beyond this specific case, both methods largely agree on the ordering of features. Sensitivity analysis reveals a more pronounced separation between the pressure feature and the remaining features, while SHAP values suggest a more gradual distribution of relevance among the inputs. 

Overall, the high degree of concordance between SHAP and gradient-based analyses provides robust evidence that the agent’s policy is primarily governed by system pressure and forecast information, with compressor-level inputs generally exerting a comparatively minor influence under the studied conditions.
\newline
\newline
\textit{Comparison to a Poorly Trained Agent} \newline \indent
To further illustrate the importance of sufficient training for meaningful policy explainability, we analyzed the SHAP attributions of an agent after only five training iterations on the 3C3F configuration. This early-stage model does not yet represent a competent control strategy, which allows us to observe how the feature relevance profile evolves before convergence.

\begin{figure}[htbp]
    \centering
    \includegraphics[width=1.0\linewidth]{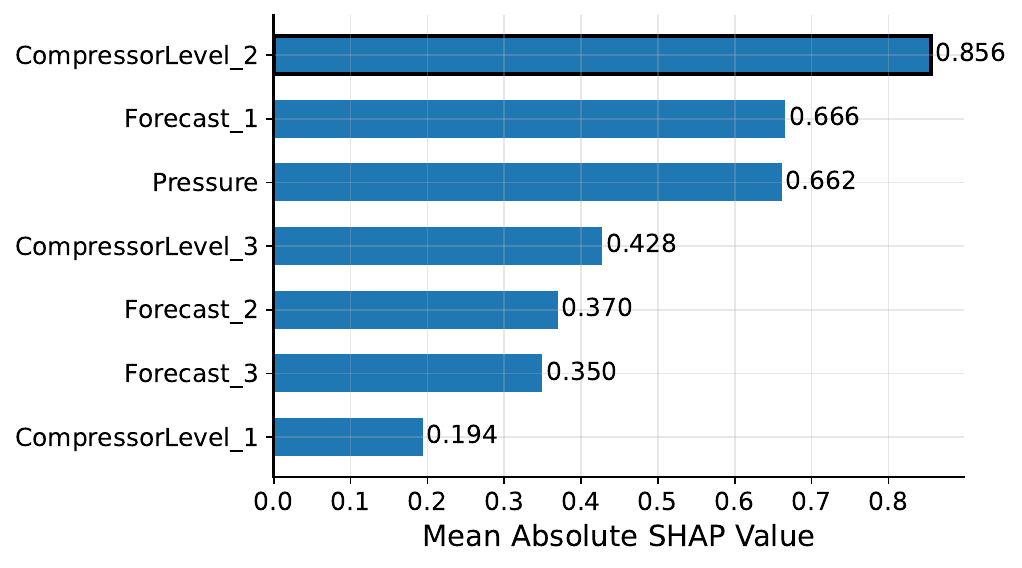}
    \caption{
        SHAP summary plot for the 3C3F configuration after only five training iterations.
        At this early checkpoint, the feature importance profile deviates markedly from the well-trained agent:
        compressor level 1 is incorrectly identified as the most relevant input, while pressure and forecast features play a subordinate role.
        This illustrates that feature attributions are only meaningful if the agent has learned plausible control logic.
    }
    \label{fig:ShapSummary3C3FCheckpoint1}
\end{figure}

Figure~\ref{fig:ShapSummary3C3FCheckpoint1} shows the mean absolute SHAP values for the agent at this preliminary checkpoint. Notably, the ordering of feature importance diverges substantially from that of the well-trained agent: here, compressor level 1 appears as the most influential input, surpassing even the pressure and first forecast features that dominate in mature policies. This indicates that, in the absence of meaningful learned behavior, the agent’s attributions may reflect arbitrary patterns, noise, or even initialization bias rather than physically plausible control logic.

Such comparisons reinforce that explainability analyses are only informative if the underlying policy has reached a reasonable level of performance and plausibility. For this reason, we present this analysis solely as an illustrative example for the 3C3F setup, rather than conducting it systematically for all configurations. This avoids overinterpreting the transient and uninformative feature relevance patterns that can arise during early or failed training phases.

\subsubsection*{Pattern-Level Attribution}

\begin{figure*}[h]
  \centering
  \begin{subfigure}[b]{0.4\textwidth}
    \includegraphics[width=\linewidth]{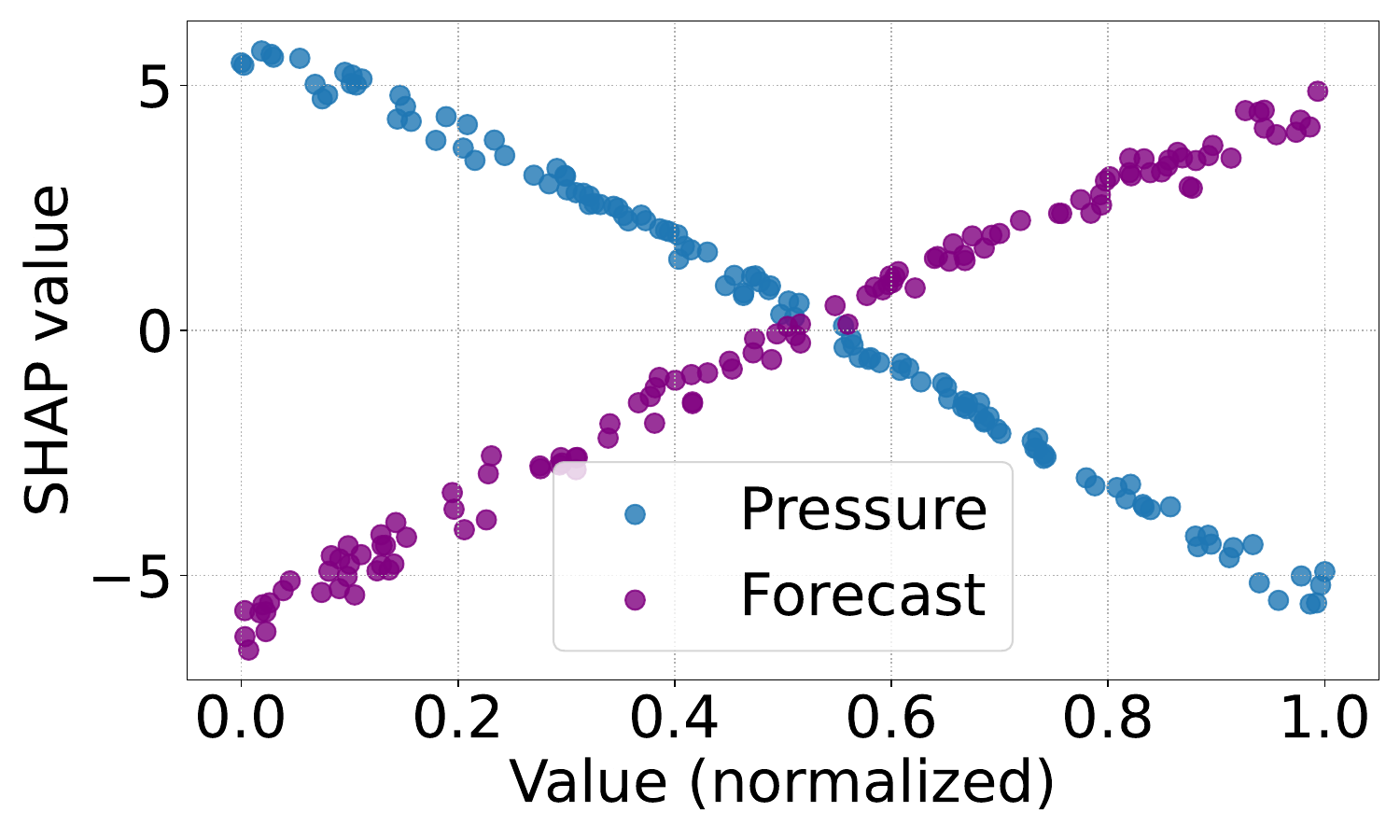}
    \caption{One step forecast $\SI{5}{s}$ and pressure}
  \end{subfigure}
  \hfill
  \begin{subfigure}[b]{0.4\textwidth}
    \includegraphics[width=\linewidth]{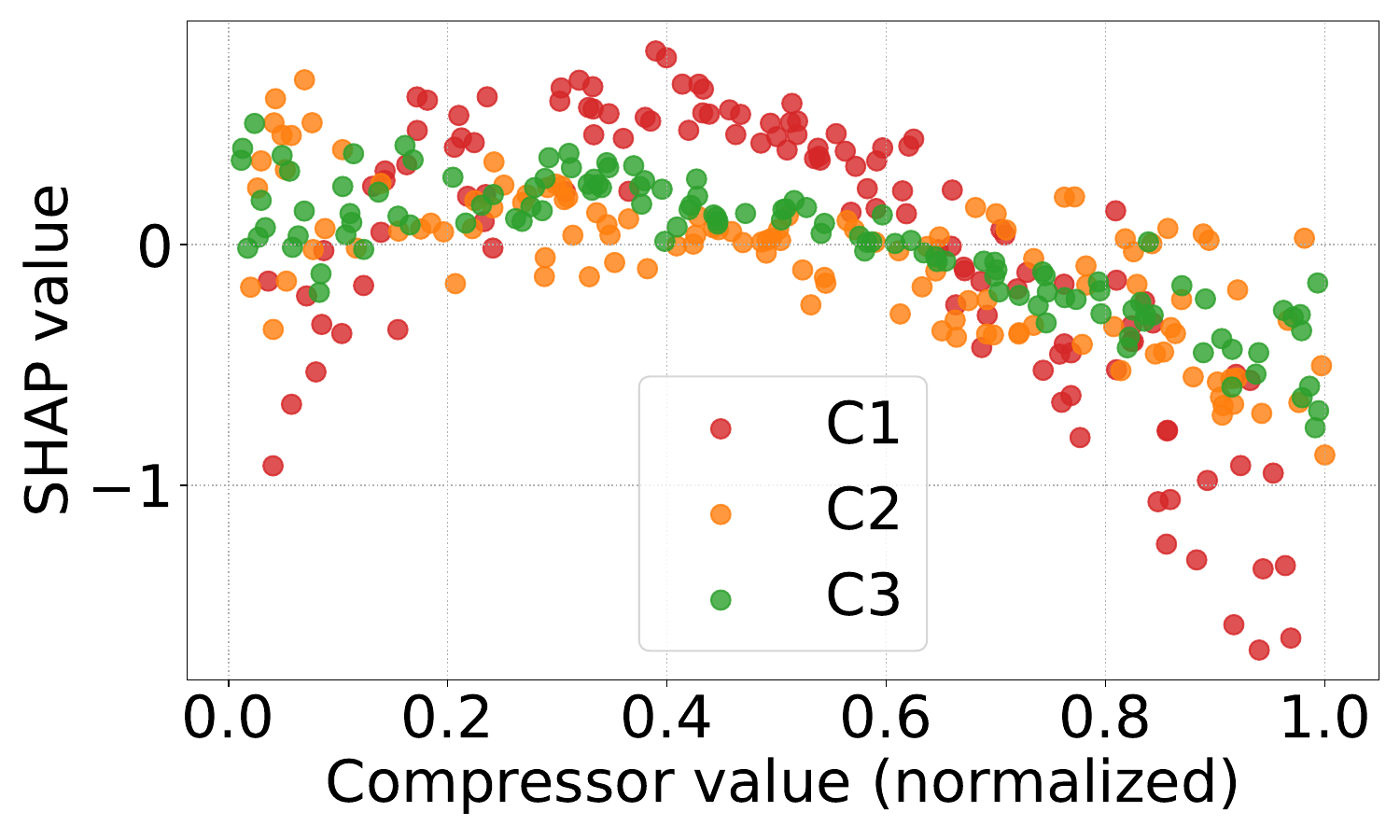}
    \caption{Compressor levels}
  \end{subfigure}
    \caption{Pattern-level attribution: SHAP scatter plots showing local contributions for selected input features in the 3C1F configuration. Each point corresponds to a single decision step.}

  \label{fig:scatter}
\end{figure*}

To further investigate the local decision behavior of the trained agent, SHAP scatter plots were generated for three feature groups in the 3C1F configuration. Each point in the scatter plots represents the SHAP value of a feature for a single decision step. For better comparability, the visualizations show the aggregated SHAP contributions per feature group. The corresponding results are shown in Figure~\ref{fig:scatter}.

The pressure shows a strong linear relationship with the SHAP values. The raw input values range approximately from minus 0.3 to plus 0.3, due to the normalization around the target pressure. Low pressure values result in large positive SHAP contributions, indicating a strong signal to increase compressor output. In contrast, high pressure values yield large negative SHAP values, as the agent learns to reduce compressor activity to avoid exceeding system limits. This behavior aligns with physical intuition and industrial control strategies.

The first forecast value exhibits a similarly strong and approximately linear correlation, but with an inverse slope. Low forecast values lead to negative SHAP values, which suggests the agent reduces compressor usage when little demand is expected. High forecast inputs close to one lead to large positive SHAP contributions, reflecting increased system output in anticipation of future demand. This indicates that the agent successfully incorporates demand forecasting into its policy.

The compressor level features show a different structure. Their SHAP values are notably smaller in magnitude, mostly between minus one and plus one, and appear noisier. Despite the noise, a weak parabolic shape is observable, with a peak around a compressor load of approximately 0.4. The fixed-speed compressor differs slightly from the variable-speed compressors and shows a more pronounced peak. This may be due to its discrete activation threshold around 50 percent, where even small variations can result in a change in output. 

At high compressor levels, the SHAP values tend to become more negative, indicating that excessive usage is penalized. At low levels, the SHAP values of the variable-speed compressors fluctuate around zero. The fixed-speed compressor, on the other hand, tends to have negative SHAP values when turned off. Although this seems counterintuitive, one possible explanation is that, in situations of low demand, even the minimum output of the fixed-speed compressor would already exceed requirements. This interpretation should be treated with caution and may require further investigation.

Figure~\ref{fig:scatter} therefore provides local interpretability for the most relevant features and reveals meaningful patterns consistent with physical understanding. The pressure and forecast inputs dominate the decision signal, while the compressor levels act as secondary modifiers with lower but structured influence.

\subsubsection*{Case-Specific Attribution}

To gain a deeper understanding of the agent’s decision-making in specific, physically relevant scenarios, SHAP waterfall plots were generated for all nine combinations of boundary and reference conditions. These cases are defined by varying the system pressure between its minimum (\(p_\text{min}\)), nominal (\(p_\text{nom}\)), and maximum (\(p_\text{max}\)) values, each combined with three levels of forecasted demand: low (0\%), medium (50\%), and high (100\%). This systematic permutation covers both operational extremes and intermediate reference points, providing a comprehensive view of policy behavior across the most relevant state space regions.

\begin{figure*}[h]
    \centering

    \begin{subfigure}[b]{0.31\textwidth}
        \includegraphics[width=\textwidth]{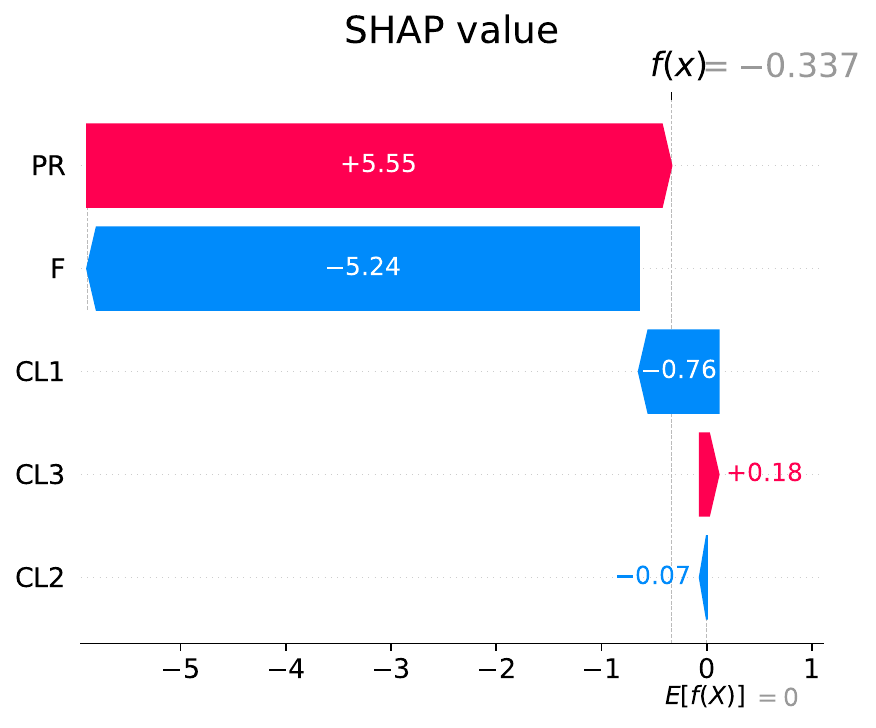}
        \caption{\(p_\text{min}\), 0\% forecast}
    \end{subfigure}
    \hfill
    \begin{subfigure}[b]{0.31\textwidth}
        \includegraphics[width=\textwidth]{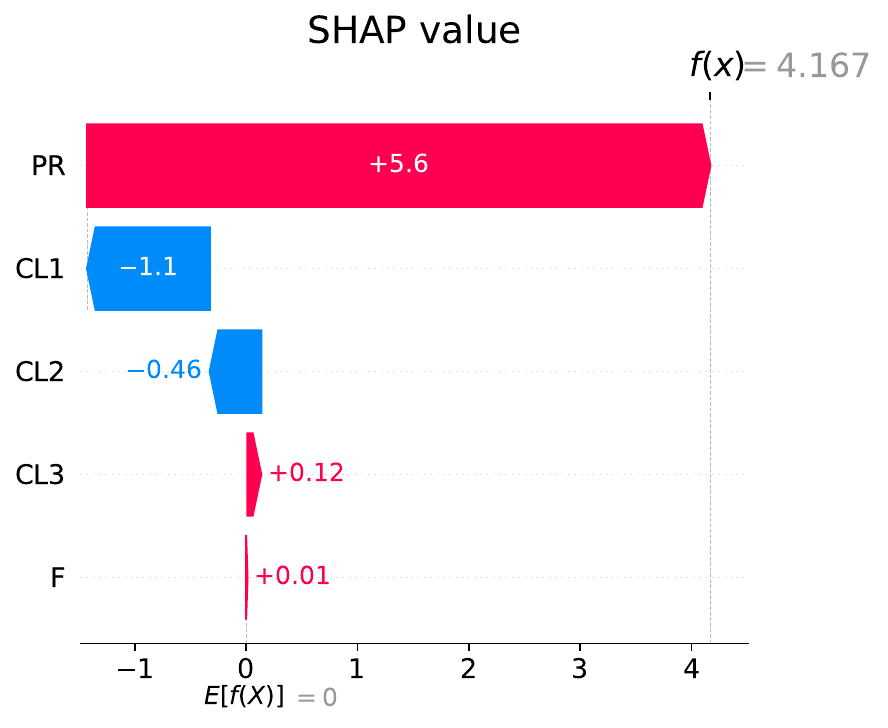}
        \caption{\(p_\text{min}\), 50\% forecast}
    \end{subfigure}
    \hfill
    \begin{subfigure}[b]{0.31\textwidth}
        \includegraphics[width=\textwidth]{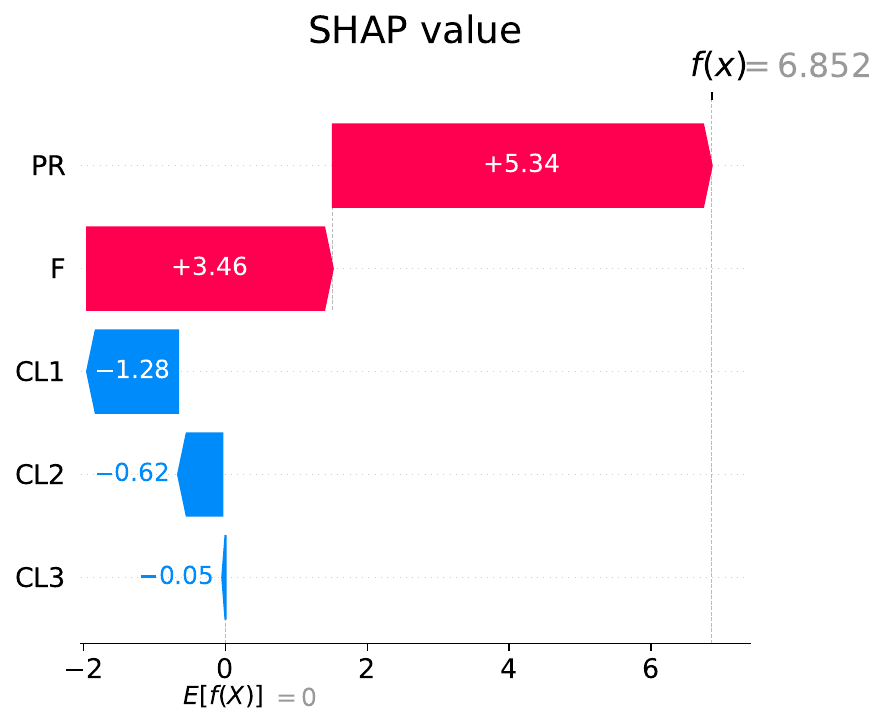}
        \caption{\(p_\text{min}\), 100\% forecast}
    \end{subfigure}

    \vspace{0.3cm}

    \begin{subfigure}[b]{0.31\textwidth}
        \includegraphics[width=\textwidth]{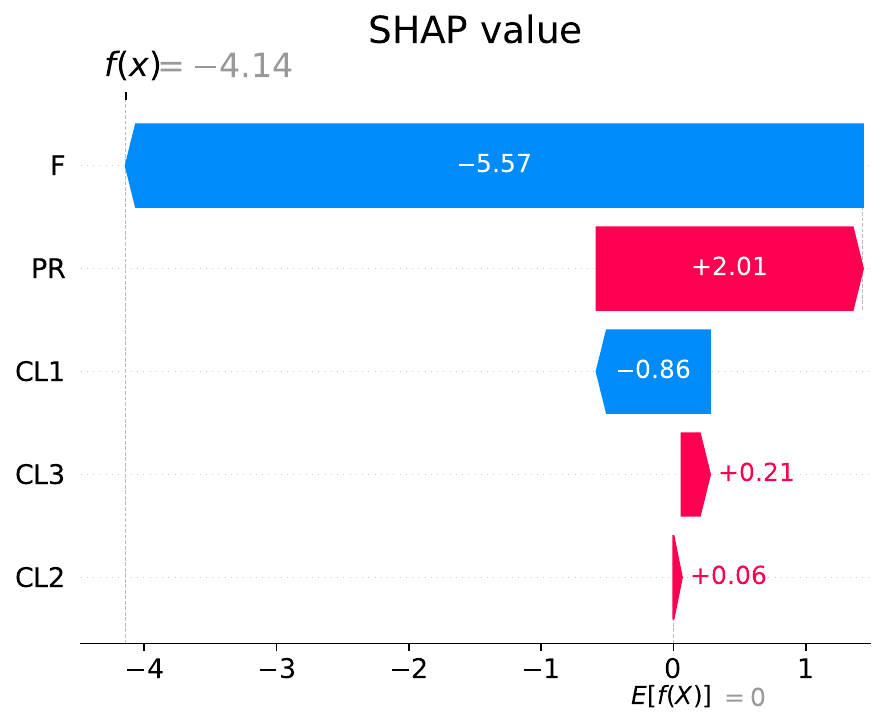}
        \caption{\(p_\text{nom}\), 0\% forecast}
    \end{subfigure}
    \hfill
    \begin{subfigure}[b]{0.31\textwidth}
        \includegraphics[width=\textwidth]{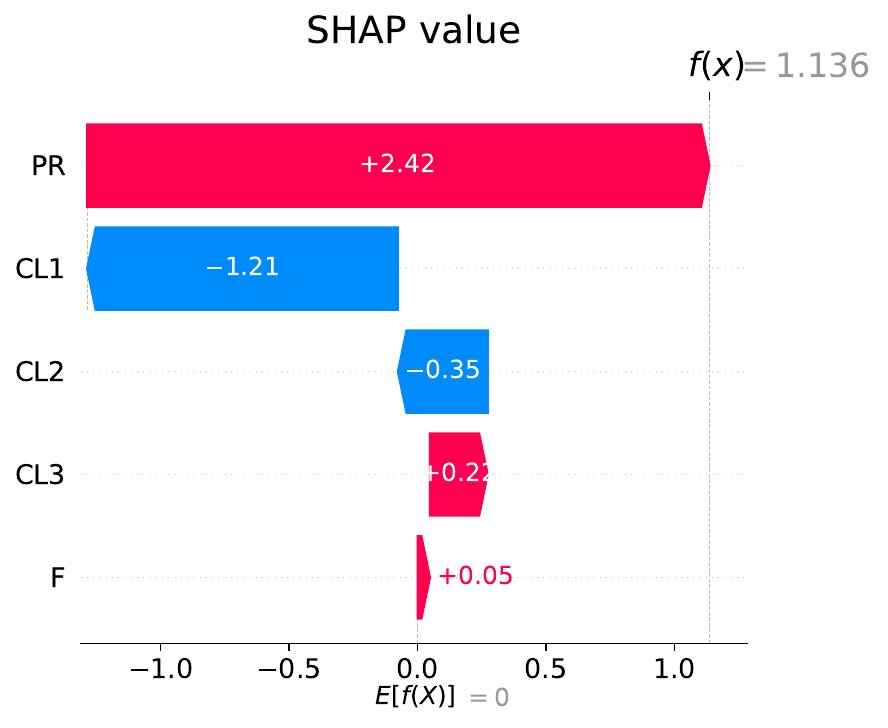}
        \caption{\(p_\text{nom}\), 50\% forecast}
    \end{subfigure}
    \hfill
    \begin{subfigure}[b]{0.31\textwidth}
        \includegraphics[width=\textwidth]{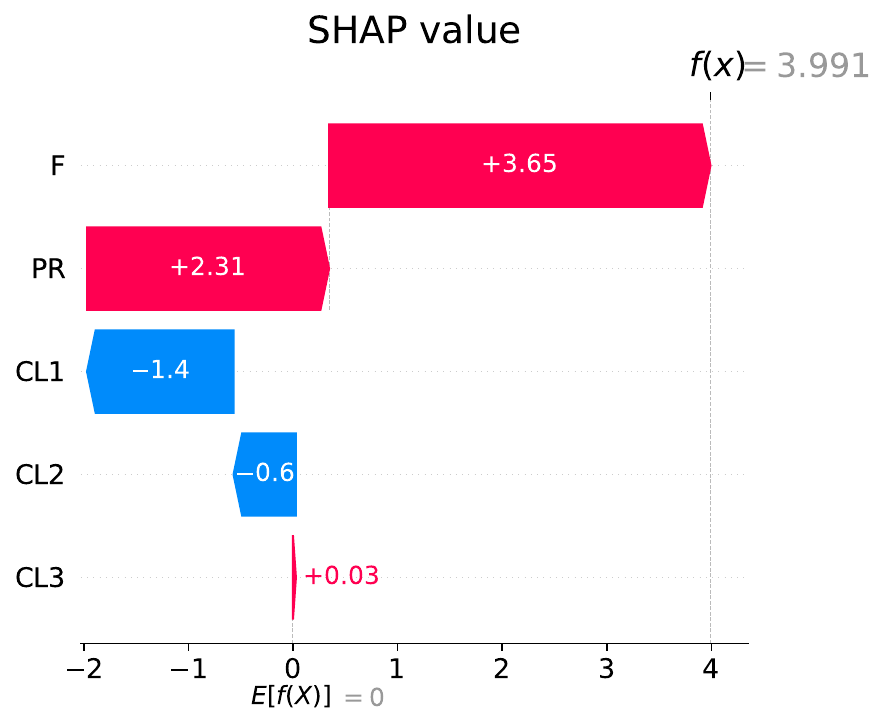}
        \caption{\(p_\text{nom}\), 100\% forecast}
    \end{subfigure}

    \vspace{0.3cm}

    \begin{subfigure}[b]{0.31\textwidth}
        \includegraphics[width=\textwidth]{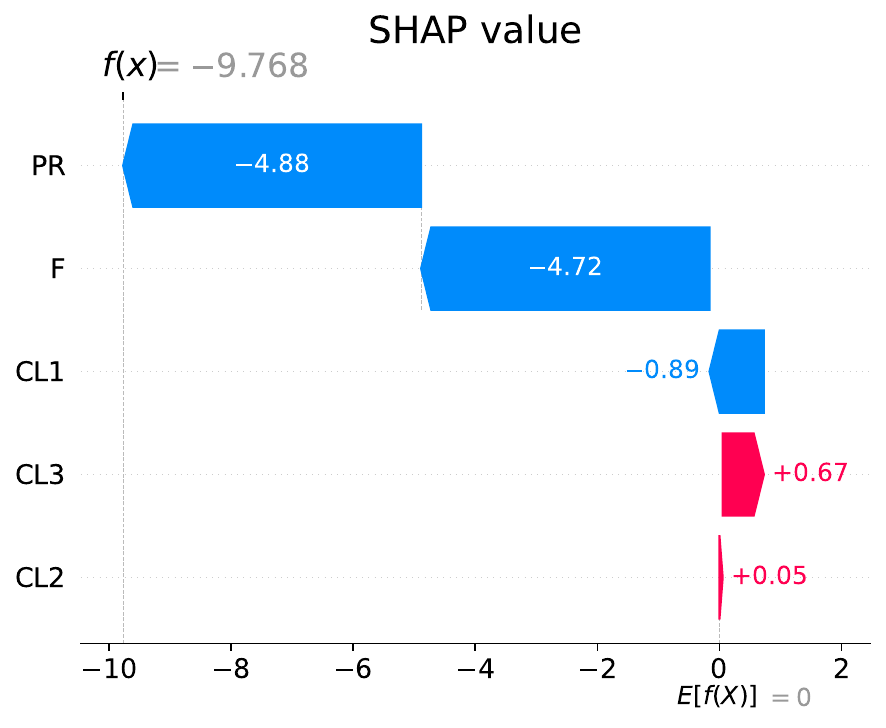}
        \caption{\(p_\text{max}\), 0\% forecast}
    \end{subfigure}
    \hfill
    \begin{subfigure}[b]{0.31\textwidth}
        \includegraphics[width=\textwidth]{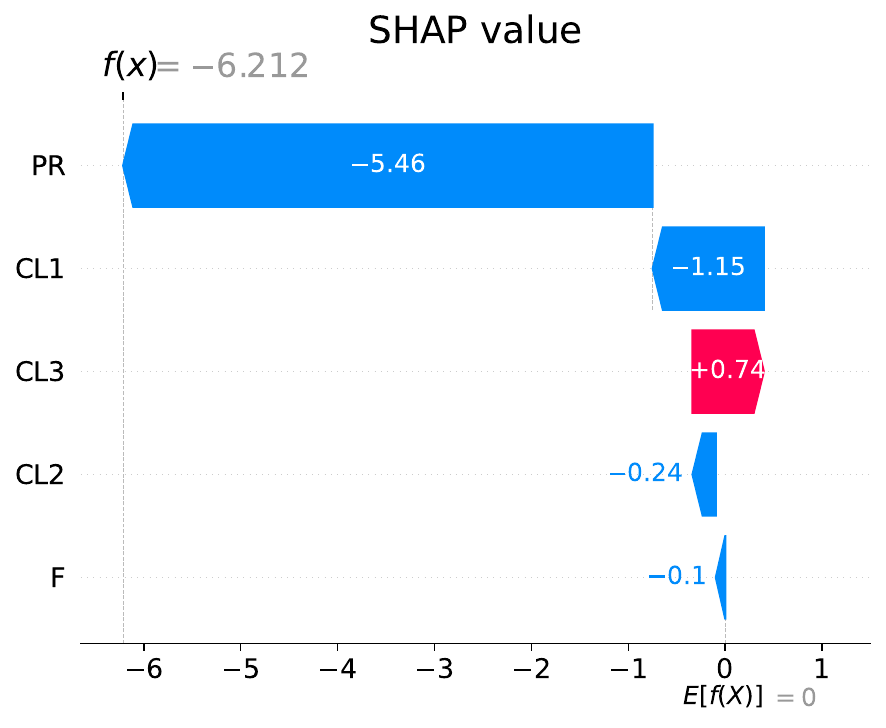}
        \caption{\(p_\text{max}\), 50\% forecast}
    \end{subfigure}
    \hfill
    \begin{subfigure}[b]{0.31\textwidth}
        \includegraphics[width=\textwidth]{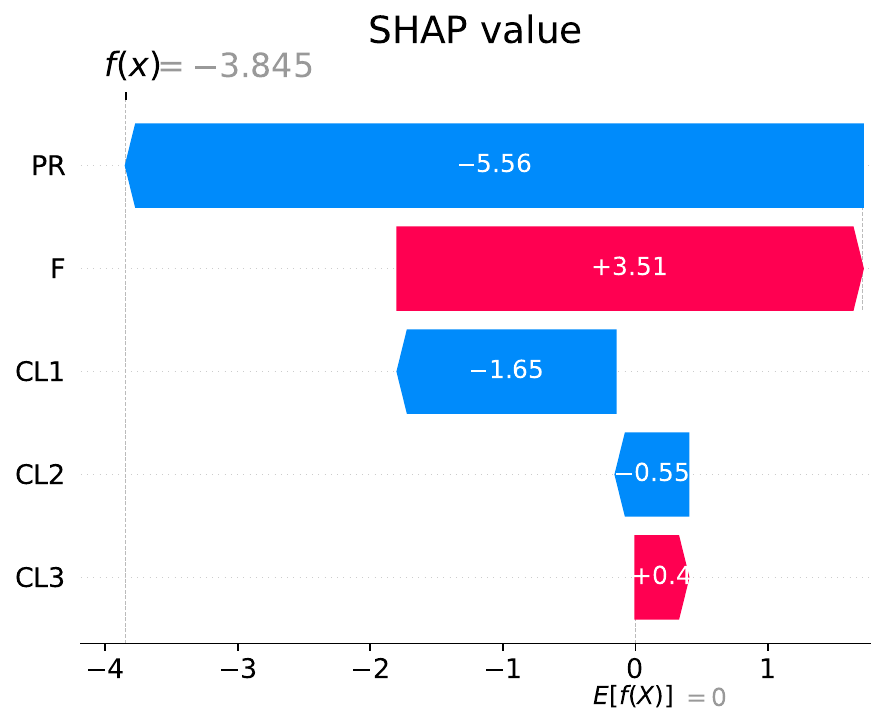}
        \caption{\(p_\text{max}\), 100\% forecast}
    \end{subfigure}

    \caption{
        Case-specific attribution: SHAP waterfall plots for all nine combinations of pressure and forecast levels.
        Rows correspond to pressure values (\(p_\text{min}\), \(p_\text{nom}\), \(p_\text{max}\)), columns to forecast values (0\%, 50\%, 100\%). 
        Feature abbreviations: PR = Pressure, F = Forecast, CL1–CL3 = Compressor Level 1–3. Each plot shows the contribution of input features to the agent's decision, starting from the expected baseline value \(E[f(X)]\).
    }
    \label{fig:SHAP_waterfall}
\end{figure*}

The results, presented in Figure~\ref{fig:SHAP_waterfall}, reveal several key insights. In all scenarios, the pressure and the first forecast value are typically the dominant contributors to the agent’s output, consistent with global and pattern-level analyses. At both the lower and upper bounds of the pressure range, the influence of the pressure input becomes especially pronounced. When the system is at its minimum pressure, positive SHAP values for pressure indicate a strong drive to increase compressor power and restore safe operating conditions. Conversely, at maximum pressure, the pressure feature exhibits large negative SHAP values, effectively suppressing compressor output to avoid overpressure situations.

A complementary effect is observed for the forecast feature. When future demand is predicted to be low (0\%), the SHAP value for the forecast is strongly negative, indicating that the agent minimizes unnecessary compressor usage in anticipation of low consumption. As the forecast increases toward 100\%, the corresponding SHAP value becomes strongly positive, resulting in proactive compressor activation to ensure sufficient supply. This behavior reflects a learned anticipation of future demand, a crucial aspect for energy-efficient control.

The influence of compressor level features remains generally modest across all scenarios. Their SHAP values typically fall within a narrow band around zero, reflecting their secondary role in decision-making compared to pressure and forecast inputs. Some variation may occur at the nominal operating point or when neither pressure nor forecast provides a decisive signal, but even then, the contribution of compressor levels remains relatively small.

By including all nine combinations, this analysis provides a detailed and robust validation of the learned policy’s physical plausibility and operational safety. The results confirm that the agent internalizes system constraints and demand anticipation, while appropriately de-emphasizing less informative features. This level of interpretability supports the argument for real-world deployment in safety-critical industrial environments.

\subsubsection*{Time-Resolved SHAP Attribution}

To explicitly relate feature attributions to the learned control behavior, SHAP values are analyzed along representative closed-loop control trajectories. Two synthetic excitation scenarios are considered: a demand sweep under constant pressure and a pressure sweep under constant demand. These scenarios are designed to isolate the agent’s response to systematic variations of individual inputs while keeping all remaining features fixed.

\begin{figure*}[h]
  \centering
  \begin{subfigure}[b]{0.48\textwidth}
    \includegraphics[width=\linewidth]{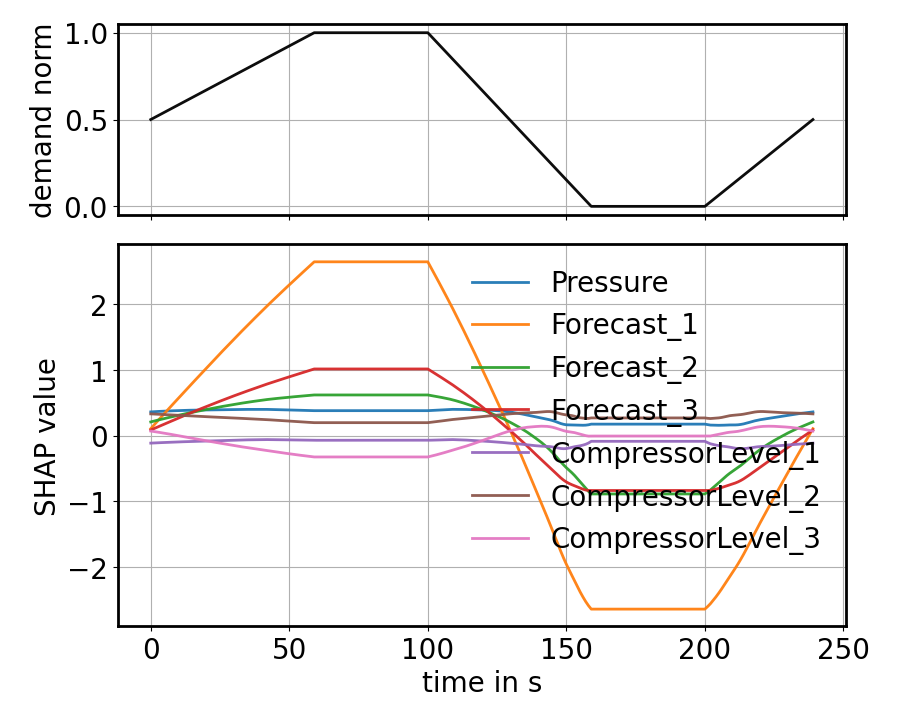}
    \caption{Time-resolved SHAP attribution for a demand sweep under constant pressure. The upper panel shows the applied demand excitation signal, while the lower panel depicts the corresponding local SHAP values over time.}
  \end{subfigure}
  \hfill
  \begin{subfigure}[b]{0.48\textwidth}
    \includegraphics[width=\linewidth]{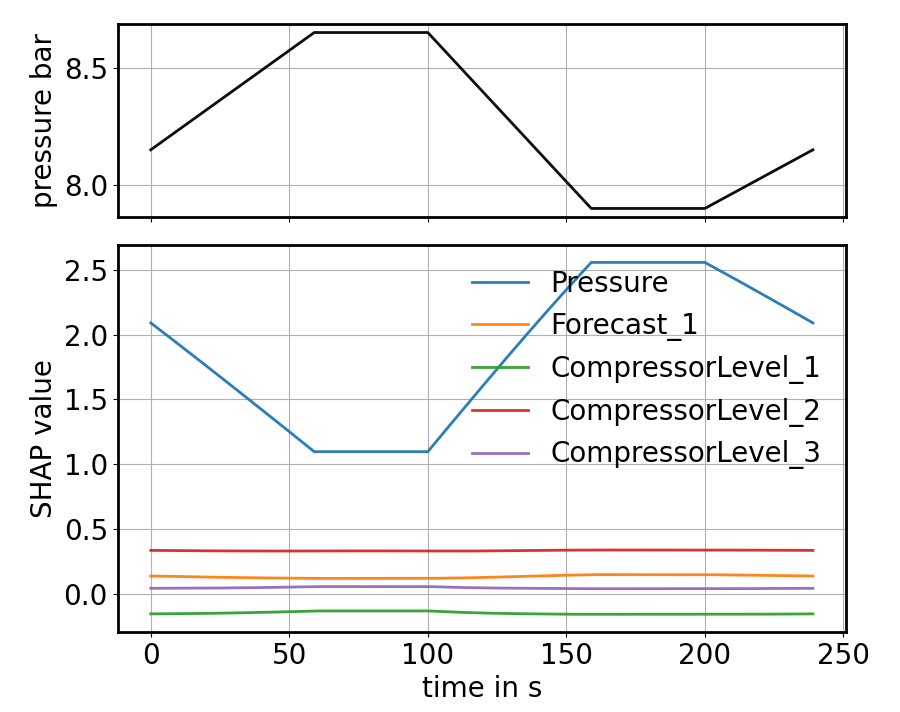}
    \caption{Time-resolved SHAP attribution for a pressure sweep under constant demand. The upper panel illustrates the pressure excitation profile, and the lower panel shows the resulting temporal evolution of SHAP values.}
  \end{subfigure}
  \caption{Time-resolved SHAP attribution aligned with representative closed-loop control trajectories. Feature attributions are evaluated locally at each time step and visualized alongside the corresponding excitation signals, enabling direct interpretation of dynamic control behavior.}
  \label{fig:timeShap}
\end{figure*}

Figure~\ref{fig:timeShap} illustrates the resulting time-resolved SHAP attributions together with the corresponding excitation signals. In the constant-pressure scenario, the demand signal is varied over time while the pressure remains fixed. Conversely, in the constant-demand scenario, the pressure follows a predefined wave-like profile. For each time step, SHAP values are computed locally and visualized alongside the applied excitation signal.

The results show that SHAP attributions evolve coherently with the applied excitation. In the demand sweep, forecast-related features dominate the attribution during increasing demand phases, while their influence becomes negative during low-demand intervals. All remaining feature contributions are comparatively small, which is consistent with the observations from the static attribution analyses. In the pressure sweep, pressure-related features gain relevance predominantly during rising and falling pressure segments, while maintaining consistently positive SHAP values. Compressor state features remain low and stable throughout both scenarios. This behavior indicates that the agent adapts its decisions logically to time-dependent input variations and reflects physically plausible control principles.

\section{Discussion}
\label{sec:discussion}

The results indicate that the reinforcement learning agent generally follows the basic physical relationships of the compressed air system. In particular, it learns to avoid operation at the technical boundaries of the system, such as excessively low or high pressures. These pressure limits are reflected not only in the agent’s direct control behavior but also in the SHAP attribution results, where extreme pressure values lead to strong corrective actions. This suggests that the agent has internalized technical safety constraints through interaction with the environment, despite not being explicitly programmed with hard limits. Moreover, the strong influence of forecast values on the agent’s decisions indicates a predictive control strategy. The agent adjusts its output not only based on the current system state but also in anticipation of future demand. This predictive behavior is evident in both the scatter and waterfall SHAP analyses and underlines the model's ability to align with industrial control logic.

Furthermore, the agent demonstrates efficiency-aware behavior. At higher system pressures, where producing additional compressed air is energetically more expensive, the agent tends to reduce compressor output. This indicates that the policy has learned to implicitly factor in energetic cost structures. While compressor-level features show relatively low SHAP relevance compared to pressure and forecast, this does not necessarily imply that they are unimportant. Rather, it may suggest that the agent prioritizes higher-level state information (such as pressure and demand) before fine-tuning individual compressor actions. An alternative explanation is that the impact of specific compressor configurations on overall efficiency is less pronounced in the given setup. Future work should investigate the role of compressor-specific actions in more detail, particularly in scenarios where switching costs, maintenance, or heterogeneous compressor types play a stronger role. In summary, the learned policy appears physically plausible, operationally safe, and efficiency-oriented, making it a promising candidate for real-world deployment.

\subsection{Outlook and Future Work}
\label{sec:outlook_and_future_work}

While the current study provides a promising foundation, several avenues remain for further research. First, a more detailed investigation into the role of compressor-level decisions is warranted. Although their SHAP contributions were relatively small, it remains unclear whether this is due to their genuinely limited impact or the agent's prioritization of pressure and forecast features. Follow-up studies could isolate the compressors' influence under varying boundary conditions, cost models, or switching penalties to evaluate their strategic relevance in more complex control settings.

A further promising research direction is the inclusion of additional contextual variables that act as proxies for latent demand drivers. In the building control domain, occupancy-related information has been shown to significantly improve control performance by enabling better anticipation of future loads and reducing energy consumption \citep{DAI2020110159}. Similar concepts could be transferred to industrial compressed air systems by incorporating production-related context, such as shift schedules, machine utilization, or calendar information, as indirect predictors of air demand. Exploring such context-aware control strategies represents an interesting avenue for future work.

Second, the generalizability of the proposed methodology to other industrial energy systems should be explored. The explainability-driven evaluation framework, especially the SHAP-based multi-level attribution and input perturbation testing, offers a structured way to assess control plausibility and operational safety. Applying this framework to systems such as fuel cells, combined heat and power (CHP) plants, or thermal storage units could reveal whether similar interpretability and trustworthiness can be achieved across domains.

Finally, real-world deployment scenarios should be addressed. This includes the integration of domain-specific energy cost models, uncertainty handling (e.g., for forecasts), and the incorporation of runtime constraints such as maintenance, degradation, or failure risk. These extensions would bring the agent closer to operational readiness while also enabling a more rigorous evaluation of trade-offs between efficiency, robustness, and system safety.

\section{Conclusion}
\label{sec:conclusion}

This paper presented an interpretable reinforcement learning approach for controlling compressed air systems in industrial environments. The proposed framework combines policy training with structured evaluation via input perturbation testing and multi-level SHAP-based feature attribution, complemented by time-resolved SHAP analyses. The results show that the agent not only respects physical and technical constraints, such as pressure boundaries, but also anticipates future demand through forecast-driven control. The policy behaves efficiently by implicitly minimizing compressor usage under high-pressure conditions, indicating a learned understanding of energetic cost structures.

Explainability analyses confirm that the agent's decisions are primarily influenced by pressure and forecast features, while compressor-level signals play a secondary role in the current setup. This prioritization aligns with intuitive control strategies and suggests that the agent internalizes high-level system dynamics before fine-tuning actuator commands. The overall behavior is physically plausible, operationally safe, and energetically reasonable, making the approach suitable for real-world deployment in industrial control scenarios.

The presented methodology serves as a blueprint for safe, interpretable, and efficient reinforcement learning in energy-critical applications and opens up future research directions for broader transfer and system-specific refinements.



\appendix

\section{Mathematical Details of Explainability Methods}
\label{app:explainability_math}

\subsection{Gradient-Based Sensitivity}
Given a policy $\pi_{\theta}(s)$ and input state \(s \in \mathbb{R}^{n_s}\), the input sensitivity is:

\[
g_j(s) = \left| \frac{\partial \pi_\theta(s)}{\partial s_j} \right|, \quad j = 1,\dots,n_s.
\]
The relevance is computed as:

\[
G_j = \frac{1}{N} \sum_{i=1}^{N} g_j(s_i).
\]

\subsection{SHAP Attribution}
SHAP values $\phi_j$ represent each feature's contribution to the output. For an input $s$, the value for feature $j$ is:

\[
\phi_j = \sum_{S \subseteq F \setminus \{j\}} \frac{|S|!(|F| - |S| - 1)!}{|F|!} \left[f_{S \cup \{j\}}(s) - f_S(s)\right].
\]
Multi-output attributions are aggregated per feature:
\[
\phi_j^\text{(total)} = \sum_k \phi_{j,k}.
\]

\section{Hyperparameters}
The hyperparameters are listed in Table~\ref{tab:hyperparameters}.
\begin{table}[htbp]
\centering
\caption{Final Selected Network Architecture and Hyperparameters}
\label{tab:hyperparameters}
\begin{tabular}{l l}
\hline
\textbf{Parameter}                        & \textbf{Selected Value}    \\ \hline
Network architecture                      & Fully connected + LSTM     \\
Neurons in first layer                    & 128                        \\
Neurons in second layer                   & 128                        \\
Use of LSTM                               & True                       \\
Learning rate                             & 0.0001                     \\
Discount factor ($\gamma$)                & 0.995                      \\
GAE Lambda ($\lambda$)                    & 0.97                       \\
Clip parameter (PPO)                      & 0.3                        \\
KL divergence coefficient                 & 0.2                        \\
Entropy coefficient                       & 0.0                        \\
Value function loss coefficient           & 1.0                        \\
Gradient clipping                         & None                       \\ \hline
\end{tabular}
\end{table}
\section*{Acknowledgments}

\subsection*{Funding}
The authors gratefully acknowledge the financial support of Fraunhofer IPA.

\subsection*{Declaration of generative AI and AI-assisted technologies in the writing process}
During the preparation of this work the authors used ChatGPT for proofreading and style improvement. 
After using this tool/service, the authors reviewed and edited the content as needed and take full responsibility for the content of the publication.

\printcredits

\bibliographystyle{cas-model2-names}

\bibliography{cas-refs}

\end{document}